\documentclass[twocolumn,10pt]{IEEEtran}
\usepackage[utf8]{inputenc}
\usepackage{amsmath,amssymb}
\usepackage[usenames]{color}
\usepackage{graphicx}
\usepackage{epstopdf}
\usepackage{pgf}
\usepackage{tikz}
\usepackage{cite}
\usepackage{algpseudocode}
\usepackage{algorithm}
\usepackage{subcaption}
\usepackage[keeplastbox]{flushend}

\newcommand{\norm}[1]{\left\|{#1}\right\|}

\newtheorem{theorem}{Theorem}
\newtheorem{corollary}{Corollary}
\newtheorem{lemma}{Lemma}

\DeclareMathOperator*{\argmin}{arg\,min}

\title{Stochastic Optimization from Distributed, Streaming Data in Rate-limited Networks}

\author{
\IEEEauthorblockN{Matthew~Nokleby,~\textit{Member, IEEE}, and Waheed~U.~Bajwa,~\textit{Senior Member, IEEE}} \thanks{Matthew Nokleby is with the Department of Electrical and Computer Engineering, Wayne State University, Detroit, MI 48202, USA (email: {\tt matthew.nokleby@wayne.edu}), and Waheed U. Bajwa is with the Department of Electrical and Computer Engineering, Rutgers University--New Brunswick, Piscataway, NJ 08854, USA (email: {\tt waheed.bajwa@rutgers.edu}). The work of WUB was supported in part by the NSF under grant CCF-1453073, by the ARO under grant W911NF-17-1-0546, and by the DARPA Lagrange Program under ONR/SPAWAR contract N660011824020. Some of the results reported here were presented at the $7$th IEEE International Workshop on Computational Advances in Multi-Sensor Adaptive Processing (CAMSAP'17)~\cite{NoklebyBajwa.ConfCAMSAP17}.}}

\begin{document}

\maketitle

\begin{abstract}
Motivated by machine learning applications in networks of sensors, internet-of-things (IoT) devices, and autonomous agents, we propose techniques for distributed stochastic convex learning from high-rate data streams. The setup involves a network of nodes---each one of which has a stream of data arriving at a constant rate---that solve a stochastic convex optimization problem by collaborating with each other over rate-limited communication links. To this end, we present and analyze two algorithms---termed distributed stochastic approximation mirror descent (D-SAMD) and {\em accelerated} distributed stochastic approximation mirror descent (AD-SAMD)---that are based on two stochastic variants of mirror descent and in which nodes collaborate via approximate averaging of the local, noisy subgradients using distributed consensus. Our main contributions are ($i$) bounds on the convergence rates of D-SAMD and AD-SAMD in terms of the number of nodes, network topology, and ratio of the data streaming and communication rates, and ($ii$) sufficient conditions for order-optimum convergence of these algorithms. In particular, we show that for sufficiently well-connected networks, distributed learning schemes can obtain order-optimum convergence even if the communications rate is small. Further we find that the use of accelerated methods significantly enlarges the regime in which order-optimum convergence is achieved; this is in contrast to the centralized setting, where accelerated methods usually offer only a modest improvement. Finally, we demonstrate the effectiveness of the proposed algorithms using numerical experiments.
\end{abstract}


\section{Introduction}
Machine learning at its core involves solving stochastic optimization (SO) problems of the form
\begin{equation}
\label{eqn:SO_prob}
	\min_{\mathbf{x} \in X} \psi(\mathbf{x}) \triangleq \min_{\mathbf{x} \in X} E_\xi[\phi(\mathbf{x},\xi)]
\end{equation}
to learn a ``model'' $\mathbf{x} \in X \subset \mathbb{R}^n$ that is then used for tasks such as dimensionality reduction, classification, clustering, regression, and/or prediction. A primary challenge of machine learning is to find a solution to the SO problem \eqref{eqn:SO_prob} without knowledge of the distribution $P(\xi)$. This involves finding an approximate solution to \eqref{eqn:SO_prob} using a sequence of $T$ training samples $\{\xi(t) \in \Upsilon\}_{t=1}^T$ drawn independently from the distribution $P(\xi)$, which is supported on a subset of $\Upsilon$. There are, in particular, two main categorizations of training data that, in turn, determine the types of methods that can be used to find approximate solutions to the SO problem. These are ($i$) \emph{batch} training data and ($ii$) \emph{streaming} training data.

In the case of {\em batch} training data, where all $T$ samples $\{\xi(t)\}$ are pre-stored and simultaneously available, a common strategy is {\em sample average approximation} (SAA) (also referred to as {\em empirical risk minimization} (ERM)), in which one minimizes the empirical average of the ``risk'' function $\phi(\cdot,\cdot)$ in lieu of the true expectation. In the case of {\em streaming} data, by contrast, the samples $\{\xi(t)\}$ arrive one-by-one, cannot be stored in memory for long, and should be processed as soon as possible. In this setting, {\em stochastic approximation} (SA) methods---the most well known of which is stochastic gradient descent (SGD)---are more common. Both SAA and SA have a long history in the literature; see~\cite{Kushner.Book2010} for a historical survey of SA methods, \cite{Kim.etal.HSO2015} for a comparative review of SAA and SA techniques, and \cite{Pereyra.etal.JSTSP16} for a recent survey of SO techniques.

Among other trends, the rapid proliferation of sensing and wearable devices, the emergence of the internet-of-things (IoT), and the storage of data across geographically-distributed data centers have spurred a renewed interest in development and analysis of new methods for learning from {\em fast-streaming} and {\em distributed} data. The goal of this paper is to find a fast and efficient solution to the SO problem \eqref{eqn:SO_prob} in this setting of distributed, streaming data. In particular, we focus on geographically-distributed nodes that collaborate over {\em rate-limited} communication links (e.g., wireless links within an IoT infrastructure) and obtain independent streams of training data arriving at a constant rate.

The relationship between the rate at which communication takes place between nodes and the rate at which streaming data arrive at individual nodes plays a critical role in this setting. If, for example, data samples arrive much faster than nodes can communicate among themselves, it is difficult for the nodes to exchange enough information to enable an SA iteration on existing data in the network before new data arrives, thereby overwhelming the network. In order to address the challenge of distributed SO in the presence of a mismatch between the communications and streaming rates, we propose and analyze two distributed SA techniques, each based on distributed averaging consensus and stochastic mirror descent. In particular, we present bounds on the convergence rates of these techniques and derive conditions---involving the number of nodes, network topology, the streaming rate, and the communications rate---under which our solutions achieve order-optimum convergence speed.

\subsection{Relationship to Prior Work}
SA methods date back to the seminal work of Robbins and Monro~\cite{Robbins.Monro.AMS51}, and recent work shows that, for stochastic convex optimization, SA methods can outperform SAA methods~\cite{Nemirovski.etal.JOO09,Juditsky.etal.SS11}. Lan~\cite{Lan.MP12} proposed {\em accelerated stochastic mirror descent}, which achieves the best possible convergence rate for general stochastic convex problems. This method, which makes use of noisy subgradients of $\psi(\cdot)$ computed using incoming training samples, satisfies
\begin{equation}\label{eqn:smd.rate}
	E[\psi(\mathbf{x}(T)) - \psi(\mathbf{x}^*)] \leq O(1)\left[\frac{L}{T^2} + \frac{\mathcal{M}+\sigma}{\sqrt{T}} \right],
\end{equation}
where $\mathbf{x}^*$ denotes the minimizer of \eqref{eqn:SO_prob}, $\sigma^2$ denotes variance of the subgradient noise, and $\mathcal{M}$ and $L$ denote the Lipschitz constants associated with the non-smooth (convex) component of $\psi$ and the gradient of the smooth (convex) component of $\psi$, respectively. Further assumptions such as smoothness and strong convexity of $\psi(\cdot)$ and/or presence of a structured regularizer term in $\psi(\cdot)$ can remove the dependence of the convergence rate on $\mathcal{M}$ and/or improve the convergence rate to $O(\sigma/T)$ \cite{Hu.etal.NIPS09,Nemirovski.etal.JOO09,Xiao.JMLR10,Chen.etal.NIPS12}.

The problem of {\em distributed} SO goes back to the seminal work of Tsitsiklis et al.~\cite{Tsitsiklis.etal.ITAC1986}, which presents distributed first-order methods for SO and gives proofs of their asymptotic convergence. Myriad works since then have applied these ideas to other settings, each with different assumptions about the type of data, how the data are distributed across the network, and how distributed units process data and share information among themselves. In order to put our work in context, we review a representative sample of these works. A recent line of work was initiated by {\em distributed gradient descent} (DGD) \cite{Nedic.Ozdaglar.ITAC2009}, in which nodes descend using gradients of local data and collaborate via averaging consensus \cite{Dimakis.etal:IEEE2010}. More recent works incorporate accelerated methods, time-varying or directed graphs, data structure, etc. \cite{Srivastava.Nedic.IJSTSP2011,Tsianos.etal.Conf2012,Mokhtari.Ribeiro.JMLR2016,Bijra.etal.arxiv2016,LiChenEtAl.N16}. These works tend not to address the SA problem directly; instead, they suppose a linearly separable function consistent with SAA using local, independent and identically distributed (i.i.d.) data. The works \cite{Ram.etal.JOTA2010,Duchi.etal.ITAC2012,RaginskyBouvrie.ConfCDC12,DuchiAgarwalEtAl.SJO12} do consider SA directly, but suppose that nodes engage in a single round of message passing per stochastic subgradient sample.

We conclude by discussing two lines of works \cite{Dekel.etal.JMLR2012,Rabbat.ConfCAMSAP15,tsianos.rabbat.SIPN2016} that are most closely related to this work. In \cite{Dekel.etal.JMLR2012}, nodes perform distributed SA by forming distributed mini-batch averages of stochastic gradients and using stochastic dual averaging.
The main assumption in this work is that nodes can compute {\em exact} stochastic gradient averages (e.g., via {\tt AllReduce} in parallel computing architectures). Under this assumption, it is shown in this work that there is an appropriate mini-batch size for which the nodes' iterates converge at the optimum (centralized) rate. However, the need for exact averages in this work is not suited to rate-limited (e.g., wireless) networks, in which mimicking the {\tt AllReduce} functionality can be costly and challenging.

The need for exact stochastic gradient averages in~\cite{Dekel.etal.JMLR2012} has been relaxed recently in~\cite{tsianos.rabbat.SIPN2016}, in which nodes carry out distributed stochastic dual averaging by computing {\em approximate} mini-batch averages of dual variables via distributed consensus. In addition, and similar to our work,~\cite{tsianos.rabbat.SIPN2016} allows for a mismatch between the communications rate and the data streaming rate. Nonetheless, there are four main distinctions between \cite{tsianos.rabbat.SIPN2016} and our work.
First, we provide results for  stochastic {\em composite} optimization, whereas \cite{Dekel.etal.JMLR2012,tsianos.rabbat.SIPN2016} suppose a differentiable objective. Second, we consider distributed {\em mirror descent}, which allows for a limited generalization to non-Euclidean settings. Third, we explicitly examine the impact of slow communications rate on performance, in particular highlighting the need for large mini-batches and their impact on convergence speed when the communications rate is slow. In \cite{tsianos.rabbat.SIPN2016}, the optimum mini-batch size is first derived from \cite{Dekel.etal.JMLR2012}, after which the communications rate needed to facilitate distributed consensus at the optimum mini-batch size is specified. While it appears to be possible to derive some of our results from a re-framing of the results of \cite{tsianos.rabbat.SIPN2016}, it is crucial to highlight the trade-offs necessary under slow communications, which is not done in prior works. Finally, this work also presents a distributed {\em accelerated} mirror descent approach to distributed SA; a somewhat surprising outcome is that acceleration substantially improves the convergence rate in networks with slow communications.

\subsection{Our Contributions}
In this paper, we present two strategies for distributed SA over networks with fast streaming data and slow communications links: distributed stochastic approximation mirror descent (D-SAMD) and accelerated distributed stochastic approximation mirror descent (AD-SAMD). In both cases, nodes first locally compute mini-batch stochastic subgradient averages to accommodate a fast streaming rate (or, equivalently, a slow communications rate), and then they collaboratively compute approximate network subgradient averages via distributed consensus. Finally, nodes individually employ mirror descent and accelerated mirror descent, respectively, on the approximate averaged subgradients for the next set of iterates.

Our main theoretical contribution is the derivation of \textit{upper} bounds on the convergence rates of D-SAMD and AD-SAMD. These bounds involve a careful analysis of the impact of imperfect subgradient averaging on individual nodes' iterates. In addition, we derive sufficient conditions for order-optimum convergence of D-SAMD and AD-SAMD in terms of the streaming and communications rates, the size and topology of the network, and the data statistics.

Two key findings of this paper are that distributed methods can achieve order-optimum convergence with small communication rates, as long as the number of nodes in the network does not grow too quickly as a function of the number of data samples each node processes, and that accelerated methods seem to offer order-optimum convergence in a larger regime than D-SAMD, thus potentially accommodating slower communications links relative to the streaming rate. By contrast, the convergence speeds of {\em centralized} stochastic mirror descent and accelerated stochastic mirror descent typically differ only in higher-order terms. We hasten to point out that we do {\em not} claim superior performance of D-SAMD and AD-SAMD versus other distributed methods. Instead, the larger goal is to establish the existence of methods for order-optimum stochastic learning in the fast-streaming, rate-limited regimes. D-SAMD and AD-SAMD should be best regarded as a proof of concept towards this end.

\subsection{Notation and Organization}
We typically use boldfaced lowercase and boldfaced capital letters (e.g., $\mathbf{x}$ and $\mathbf{W}$) to denote (possibly random) vectors and matrices, respectively. Unless otherwise specified, all vectors are assumed to be column vectors. We use $(\cdot)^T$ to denote the transpose operation and $\mathbf{1}$ to denote the vector of all ones. Further, we denote the expectation operation by $E[\cdot]$ and the field of real numbers by $\mathbb{R}$. We use $\nabla$ to denote the gradient operator, while $\odot$ denotes the Hadamard product. Finally, given two functions $p(r)$ and $q(r)$, we write $p(r) = O(q(r))$ if there exists a constant $C$ such that $\forall r, p(r) \leq C q(r)$, and we write $p(r) = \Omega(q(r))$ if $q(r) = O(p(r))$.

The rest of this paper is organized as follows. In Section~\ref{sect:setting}, we formalize the problem of distributed stochastic composite optimization. In Sections~\ref{sect:mirror.descent} and \ref{sect:accelerated.mirror.descent}, we describe D-SAMD and AD-SAMD, respectively, and also derive performance guarantees for these two methods. We examine the empirical performance of the proposed methods via numerical experiments in Section~\ref{sect:numerical}, and we conclude the paper in Section~\ref{sect:conclusion}. Proofs are provided in the appendix.

\section{Problem Formulation}\label{sect:setting}
The objective of this paper is order-optimal, distributed minimization of the composite function
\begin{equation}
	\psi(\mathbf{x}) = f(\mathbf{x}) + h(\mathbf{x}),
\end{equation}
where $\mathbf{x} \in X \subset \mathbb{R}^n$ and $X$ is convex and compact. The space $\mathbb{R}^n$ is endowed with an inner product $\langle \cdot , \cdot \rangle$ that need not be the usual one and a norm $\norm{\cdot}$ that need not be the one induced by the inner product. In the following, the minimizer and the minimum value of $\psi$ are denoted as:
\begin{equation}
	\mathbf{x}^* \triangleq \arg\min_{\mathbf{x} \in X} \psi(\mathbf{x}), \quad \text{and} \quad \psi^* \triangleq \psi(\mathbf{x}^*).
\end{equation}

We now make a few assumptions on the smooth ($f(\cdot)$) and non-smooth ($h(\cdot)$) components of $\psi$. The function $f: X \to \mathbb{R}$ is convex with Lipschitz continuous gradients, i.e.,
\begin{equation}
	\norm{\nabla f(\mathbf{x}) - \nabla f(\mathbf{y})}_* \leq L\norm{\mathbf{x} - \mathbf{x}}, \ \forall \ \mathbf{x},\mathbf{y} \in X,
\end{equation}
where $\norm{\cdot}_*$ is the dual norm associated with $\langle \cdot, \cdot \rangle$ and $\norm{\cdot}$:
\begin{equation}
	\norm{\mathbf{g}}_* \triangleq \sup_{\norm{\mathbf{x}} \leq 1} \langle \mathbf{g}, \mathbf{x} \rangle.
\end{equation}
The function $h: X \to \mathbb{R}$ is convex and Lipschitz continuous:
\begin{equation}
	\norm{h(\mathbf{x}) - h(\mathbf{y})} \leq \mathcal{M}\norm{\mathbf{x} - \mathbf{y}}, \forall \ \mathbf{x},\mathbf{y} \in X.
\end{equation}
Note that $h$ need not have gradients; however, since it is convex we can consider its {\em subdifferential}, denoted by $\partial h(\mathbf{y})$:
\begin{equation}
	\partial h(\mathbf{y}) = \{\mathbf{g}: h(\mathbf{z}) \geq h(\mathbf{y}) + \mathbf{g}^T(\mathbf{z} - \mathbf{y}), \forall \ \mathbf{z} \in X\}.
\end{equation}

An important fact that will be used in this paper is that the \emph{subgradient} $\mathbf{g} \in \partial h$ of a Lipschitz-continuous convex function $h$ is bounded~\cite[Lemma~2.6]{Shalev-Shwartz.Book2012}:
\begin{equation}
	\norm{\mathbf{g}}_* \leq \mathcal{M}, \ \forall \mathbf{g} \in \partial h(\mathbf{y}), \ \mathbf{y} \in X.
\end{equation}
Consequently, the gap between the subgradients of $\psi$ is bounded: $\forall \mathbf{x},\mathbf{y} \in X$ and $\mathbf{g}_\mathbf{x} \in \partial h(\mathbf{x})$, $\mathbf{g}_\mathbf{y} \in \partial h(\mathbf{y})$, we have
\begin{align}
	\norm{\partial \psi(\mathbf{x}) - \partial \psi(\mathbf{y})}_* &= \norm{\nabla f(\mathbf{x}) - \nabla f(\mathbf{y}) + \mathbf{g}_\mathbf{x} - \mathbf{g}_\mathbf{y}}_* \notag  \\
    &\leq \norm{\nabla f(\mathbf{x}) - \nabla f(\mathbf{y})}_* + \norm{\mathbf{g}_\mathbf{x} - \mathbf{g}_\mathbf{y}}_* \notag\\
    &\leq L\norm{\mathbf{x} - \mathbf{y}} + 2\mathcal{M}. \label{eqn:subgradient.bound}
\end{align}

\subsection{Distributed Stochastic Composite Optimization}
Our focus in this paper is minimization of $\psi(\mathbf{x})$ over a network of $m$ nodes, represented by the undirected graph $G=(V,E)$. To this end, we suppose that nodes minimize $\psi$ collaboratively by exchanging subgradient information with their neighbors at each communications round. Specifically, each node $i \in V$ transmits a message at each communications round to each of its neighbors, defined as
\begin{equation}
 \mathcal{N}_i = \{j \in V: (i,j) \in E\},
\end{equation}
where we suppose that a node is in its own neighborhood, i.e., $i \in \mathcal{N}_i$. We assume that this message passing between nodes takes place without any error or distortion. Further, we constrain the messages between nodes to be members of the dual space of $X$ and to satisfy causality; i.e., messages transmitted by a node can depend only on its local data and previous messages received from its neighbors.

Next, in terms of data generation, we suppose that each node $i \in V$ queries a first-order stochastic ``oracle'' at a fixed rate---which may be different from the rate of message exchange---to obtain noisy estimates of the subgradient of $\psi$ at different query points in $X$. Formally, we use `$t$' to index time according to {\em data-acquisition} rounds and define $\{\xi_i(t) \in \Upsilon\}_{t \geq 1}$ to be a sequence of independent (with respect to $i$ and $t$) and identically distributed (i.i.d.) random variables with unknown probability distribution $P(\xi)$. At each data-acquisition round $t$, node $i$ queries the oracle at search point $\mathbf{x}_i(s)$ to obtain a point $G(\mathbf{x}_i(s),\xi_i(t))$ that is a noisy version of the subgradient of $\psi$ at $\mathbf{x}_i(s)$. Here, we use `$s$' to index time according to \emph{search-point update} rounds, with possibly multiple data-acquisition rounds per search-point update. The reason for allowing the search-point update index $s$ to be different from the data-acquisition index $t$ is to accommodate the setting in which data (equivalently, subgradient estimates) arrive at a much faster rate than the rate at which nodes can communicate with each other; we will elaborate further on this in the next subsection.

Formally, $G(\mathbf{x},\xi)$ is a Borel function that satisfies the following properties:
\begin{align}
	E[G(\mathbf{x},\xi)] &\triangleq \mathbf{g}(\mathbf{x}) \in  \partial \psi(\mathbf{x}), \quad \text{and}\\
    E[\norm{G(\mathbf{x},\xi) - \mathbf{g}(\mathbf{x})}_*^2] &\leq \sigma^2,
\end{align}
where the expectation is with respect to the distribution $P(\xi)$. We emphasize that this formulation is equivalent to that in which the objective function is $\psi(\mathbf{x}) \triangleq E[\phi(\mathbf{x},\xi)]$, and where nodes in the network acquire data point $\{\xi_i(t)\}_{i\in V}$ at each data-acquisition round $t$ that are then used to compute the subgradients of $\phi(\mathbf{x},\xi_i(t))$, which---in turn---are noisy subgradients of $\psi(\mathbf{x})$.

\subsection{Mini-batching for Rate-Limited Networks}
A common technique to reduce the variance of the (sub)gradient noise and/or reduce the computational burden in centralized SO is to average ``batches'' of oracle outputs into a single (sub)gradient estimate. This technique, which is referred to as \emph{mini-batching}, is also used in this paper; however, its purpose in our distributed setting is to both reduce the subgradient noise variance \emph{and} manage the potential mismatch between the communications rate and the data streaming rate. Before delving into the details of our mini-batch strategy, we present a simple model to parametrize the mismatch between the two rates. Specifically, let $\rho >0$ be the {\em communications ratio}, i.e. the fixed ratio between the rate of communications and the rate of data acquisition. That is, $\rho \geq 1$ implies nodes engage in $\rho$ rounds of message exchanges for every data-acquisition round. Similarly, $\rho < 1$ means there is one communications round for every $1/\rho$ data-acquisition rounds. We ignore rounding issues for simplicity.

The mini-batching in our distributed problem proceeds as follows. Each mini-batch round spans $b \geq 1$ data-acquisition rounds and coincides with the search-point update round, i.e., each node $i$ updates its search point at the end of a mini-batch round. In each mini-batch round $s$, each node $i$ uses its current search point $\mathbf{x}_i(s)$ to compute an average of oracle outputs
\begin{equation}
	\theta_i(s) = \frac{1}{b}\sum_{t = (s-1)b +1}^{sb} G(\mathbf{x}_i(s),\xi_i(t)).
\end{equation}
This is followed by each node computing a new search point $\mathbf{x}_i(s+1)$ using $\theta_i(s)$ and messages received from its neighbors.

In order to analyze the mini-batching distributed SA techniques proposed in this work, we need to generalize the usual averaging property of variances to non-Euclidean norms.
\begin{lemma}\label{lem:average.variance}
	Let $\mathbf{z}_1,\dots,\mathbf{z}_k$ be i.i.d. random vectors in $\mathbb{R}^n$ with $E[\mathbf{z}_i] = 0$ and $E[\norm{\mathbf{z}_i}^2_*] \leq \sigma^2$. There exists a constant $C_* \geq 0$, which depends only on $\norm{\cdot}$ and $\langle \cdot, \cdot \rangle$, such that
    \begin{equation}
    	E\left[\norm{\frac{1}{k}\sum_{i=1}^k \mathbf{z}_i  }_*^2\right] \leq \frac{C_* \sigma^2}{k}.
    \end{equation}
\end{lemma}
\begin{IEEEproof}
	This follows directly from the property of norm equivalence in finite-dimensional spaces.
\end{IEEEproof}
In order to illustrate Lemma~\ref{lem:average.variance}, notice that when $\norm{\cdot} = \norm{\cdot}_1$, i.e., the $\ell_1$ norm, and $\langle \cdot, \cdot \rangle$ is the standard inner product, the associated dual norm is the $\ell_\infty$ norm: $\norm{\cdot}_* = \norm{\cdot}_\infty$. Since $\norm{\mathbf{x}}^2_\infty \leq \norm{\mathbf{x}}^2_2 \leq n\norm{\mathbf{x}}_\infty^2$, we have $C_* = n$ in this case. Thus, depending on the norm in use, the extent to which averaging reduces subgradient noise variance may depend on the dimension of the optimization space.

In the following, we will use the notation $\mathbf{z}_i(s) \triangleq \theta_i(s) - \mathbf{g}(\mathbf{x}_i(s))$. Then, $E[\norm{\mathbf{z}_i(s)}_*^2] \leq C_*\sigma^2/b$. We emphasize that the subgradient noise vectors $\mathbf{z}_i(s)$ depend on the search points $\mathbf{x}_i(s)$; we suppress this notation for brevity.

\subsection{Problem Statement}
It is straightforward to see that mini-batching induces a performance trade-off: Averaging reduces subgradient noise and processing time, but it also reduces the rate of search-point updates (and hence slows down convergence). This trade-off depends on the relationship between the streaming and communications rates. In order to carry out distributed SA in an order-optimal manner, we will require that the nodes collaborate by carrying out $r \geq 1$ rounds of averaging consensus on their mini-batch averages $\theta_i(s)$ in each mini-batch round $s$ (see Section~\ref{sect:mirror.descent} for details). In order to complete the $r$ communication rounds in time for the next mini-batch round, we have the constraint
\begin{equation}
	r \leq b \rho.
\end{equation}
If communications is faster, or if the mini-batch rounds are longer, nodes can fit in more rounds of information exchange between each mini-batch round or, equivalently, between each search-point update. But when the mismatch factor $\rho$ is small, the mini-batch size $b$ needed to enable sufficiently many consensus rounds may be so large that the reduction in subgradient noise is outstripped by the reduction in search-point updates and the resulting convergence speed is sub-optimum. In this context, our main goal is specification of sufficient conditions for $\rho$ such that the resulting convergence speeds of the proposed distributed SA techniques are optimum.

\section{Distributed Stochastic Approximation\\Mirror Descent}\label{sect:mirror.descent}
In this section we present our first distributed SA algorithm, called \emph{distributed stochastic approximation mirror descent} (D-SAMD). This algorithm is based upon stochastic approximated mirror descent, which is a generalized version of stochastic subgradient descent. Before presenting D-SAMD, we review a few concepts that underlie mirror descent.

\begin{figure*}[t]
	\centering
	\includegraphics[width=0.95\textwidth]{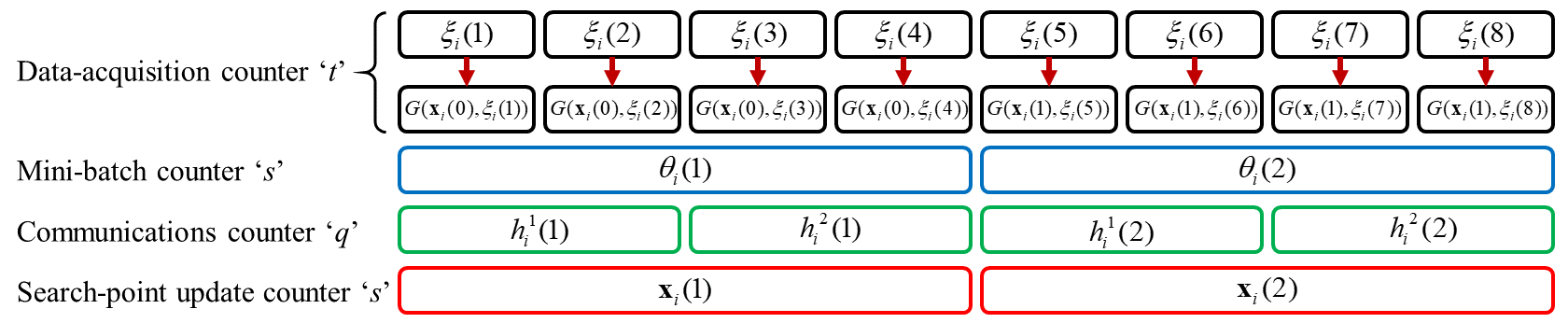}
	\caption{The different time counters for the rate-limited framework of this paper, here with $\rho = 1/2$, $b=4$, and $r=2$. In this particular case, over $T=8$ total data-acquisition rounds, each node receives $8$ data samples and computes $8$ (sub)gradients; it averages those (sub)gradients into $S=2$ mini-batch (sub)gradients; it then engages in $r=2$ rounds of consensus averaging to produce $S=2$ {\em locally averaged} (sub)gradients; each of those (sub)gradients is finally used to update the search points twice, one for each $1 \leq s \leq S$. Note that, while not explicitly shown in the figure, the search point $\mathbf{x}_i(1)$ is used for all computations spanning the data-acquisition rounds $5 \leq t \leq 8$.}\label{fig:timing_diag}
\end{figure*}

\subsection{Stochastic Mirror Descent Preliminaries}
Stochastic mirror descent, presented in \cite{Lan.MP12}, is a generalization of stochastic subgradient descent. This generalization is characterized by a {\em distance-generating function} $\omega: X \to \mathbb{R}$ that generalizes the Euclidean norm. The distance-generating function must be continuously differentiable and strongly convex with modulus $\alpha$, i.e.
\begin{equation}
	\langle \nabla \omega(\mathbf{x}) - \nabla \omega(\mathbf{y}), \mathbf{x} - \mathbf{y} \rangle \geq \alpha \norm{\mathbf{x} - \mathbf{y}}^2, \forall \ \mathbf{x},\mathbf{y} \in X.
\end{equation}
In the convergence analysis, we will require two measures of the ``radius'' of $X$ that will arise in the convergence analysis, defined as follows:
\begin{equation*}
	D_\omega \triangleq \sqrt{\max_{\mathbf{x} \in X}\omega(x) - \min_{\mathbf{x} \in X} \omega(x)}, \quad \Omega_\omega \triangleq \sqrt{\frac{2}{\alpha} D_\omega}.
\end{equation*}

The distance-generating function induces the {\em prox function}, or the Bregman divergence $V : X \times X \to \mathbb{R}_+$, which generalizes the Euclidean distance:
\begin{equation}
	V(\mathbf{x},\mathbf{z}) = \omega(\mathbf{z}) - (\omega(\mathbf{x}) + \langle \nabla \omega(\mathbf{x}), \mathbf{z}-\mathbf{x} \rangle).
\end{equation}
The prox function $V(\mathbf{x},\cdot)$ inherits strong convexity from $\omega(\cdot)$, but it need not be symmetric or satisfy the triangle inequality. We define the {\em prox mapping} $P_{\mathbf{x}}: \mathbb{R}^n \to X$ as
\begin{equation}\label{eqn:prox.mapping}
	P_\mathbf{x}(\mathbf{y}) = \argmin_{\mathbf{z} \in X} \langle \mathbf{y}, \mathbf{z}-\mathbf{x} \rangle + V(\mathbf{x},\mathbf{z}).
\end{equation}
The prox mapping generalizes the usual subgradient descent step, in which one minimizes the local linearization of the objective function regularized by the Euclidean distance of the step taken. In (centralized) stochastic mirror descent, one computes iterates of the form
\begin{align*}
	\mathbf{x}(s+1) &= P_{\mathbf{x}(s)}(\gamma_s \mathbf{g}(s))
\end{align*}
where $\mathbf{g}(s)$ is a stochastic subgradient of $\psi(\mathbf{x}(s))$, and $\gamma_s$ is a step size. These iterates have the same form as (stochastic) subgradient descent; indeed, choosing $\omega(\mathbf{x}) = \frac{1}{2}\norm{\mathbf{x}}^2_2$ as well as $\langle \cdot, \cdot \rangle$ and $\norm{\cdot}$ to be the usual ones results in subgradient descent iterations.

One can speed up convergence by choosing $\omega(\cdot)$ to match the structure of $X$ and $\psi$. For example, if the optimization space $X$ is the unit simplex over $\mathbb{R}^n$, one can choose $\omega(\mathbf{x}) = \sum_i x_i \log(x_i)$ and $\norm{\cdot}$ to be the $\ell_1$ norm. This leads to $V(\mathbf{x},\mathbf{z}) = D(\mathbf{z} || \mathbf{x}))$, where $D(\cdot || \cdot)$ denotes the Kullback-Leibler (K-L) divergence between $\mathbf{z}$ and $\mathbf{x}$. %
This choice speeds up convergence on the order of $O\left(\sqrt{n/\log(n)}\right)$ over using the Euclidean norm throughout. Along a similar vein, when $\psi$ includes an $\ell_1$ regularizer to promote a sparse minimizer, one can speed up convergence by choosing $\omega(\cdot)$ to be a $p$-norm with $p = \log(n)/(\log(n)-1)$.

In order to guarantee convergence for D-SAMD, we need to restrict further the distance-generating function $\omega(\cdot)$. In particular, we require that the resulting prox mapping be 1-Lipschitz continuous in $\mathbf{x},\mathbf{y}$ pairs, i.e., $\forall \ \mathbf{x},\mathbf{x}^\prime,\mathbf{y},\mathbf{y}^\prime \in \mathbb{R}^n$,
\begin{equation*}
	\norm{P_{\mathbf{x}}(\mathbf{y}) - P_{\mathbf{x}^\prime}(\mathbf{y}^\prime)} \leq \norm{\mathbf{x}-\mathbf{x}^\prime} + \norm{\mathbf{y} - \mathbf{y}\prime}.
\end{equation*}
This condition is in addition to the conditions one usually places on the Bregman divergence for stochastic optimization; we will use it to guarantee that imprecise gradient averages make a bounded perturbation in the iterates of stochastic mirror descent. The condition holds whenever the prox mapping is the projection of a 1-Lipschitz function of $\mathbf{x},\mathbf{y}$ onto $X$. For example, it is easy to verify that this condition holds in the Euclidean setting. One can also show that when the distance-generating function $\omega(\mathbf{x})$ is an $\ell_p$ norm for $p > 1$, the resulting prox mapping is 1-Lipschitz continuous in $\mathbf{x}$ and $\mathbf{y}$ as required.

However, not all Bregman divergences satisfy this condition. One can show that the K-L divergence results in a prox mapping that is not Lipschitz. Consequently, while we present our results in terms of general prox functions, we emphasize that the results do not apply in all cases.\footnote{We note further that it is possible to relax the constraint that the best Lipschitz constant be no larger than unity. This worsens the scaling laws---in particular, the required communications ratio $\rho$ grows in $T$ rather than decreases---and we omit this case for brevity's sake.} One can think of the results primarily in the setting of Euclidean (accelerated) stochastic subgradient descent---for which case they are guaranteed to hold---with the understanding that one can check on a case-by-case basis to see if they hold for a particular non-Euclidean setting.

\subsection{Description of D-SAMD}\label{sect:dsamd.description}
Here we present in detail D-SAMD, which generalizes stochastic mirror descent to the setting of distributed, streaming data. In D-SAMD, nodes carry out iterations similar to stochastic mirror descent as presented in \cite{Lan.MP12}, but instead of using local stochastic subgradients associated with the local search points, they carry out approximate consensus to estimate the {\em average} of stochastic subgradients across the network. This reduces the subgradient noise at each node and speeds up convergence.

Let $\mathbf{W}$ be a symmetric, doubly-stochastic matrix consistent with the network graph $G$, i.e., $[\mathbf{W}]_{ij} \triangleq w_{ij}= 0$ if $(i,j) \notin E$. Further suppose that $\mathbf{W} - \mathbf{1}\mathbf{1}^T/n$ has spectral radius strictly less than one, i.e. the second-largest eigenvalue magnitude is strictly less than one. This condition is guaranteed by choosing the diagonal elements of $\mathbf{W}$ to be strictly greater than zero.

Next, we focus on the case of constant step size $\gamma$ and set it as $0 < \gamma \leq \alpha/(2L)$.\footnote{It is shown in \cite{Lan.MP12} that a constant step size is sufficient for order-optimal performance, so we adopt such a rule here.} For simplicity, we suppose that there is a predetermined number of data-acquisition rounds $T$, which leads to $S=T/b$ mini-batch rounds. We detail the steps of D-SAMD in Algorithm \ref{alg:standard}. Further, in Figure \ref{fig:timing_diag} we illustrate the data acquisition round, mini-batch round, communication round, and search point update counters and their role in the D-SAMD algorithm.

\begin{algorithm}[t]
  \caption{Distributed stochastic approximation mirror descent (D-SAMD)
    \label{alg:standard}}
    \begin{algorithmic}[1]
    \Require Doubly-stochastic matrix $\mathbf{W}$, step size $\gamma$, number of consensus rounds $r$, batch size $b$, and stream of mini-batched subgradients $\theta_i(s)$.
    \For{$i=1:m$}
    	\State $\mathbf{x}_i(1) \gets \min_{\mathbf{x} \in X} \omega(\mathbf{x})$ \Comment{Initialize search points}
    \EndFor
    \For {$s=1:S$}
        \State $\mathbf{h}_i^0(s) \gets \theta_i(s)$ \Comment{Get mini-batched subgradients}
        \For{$q=1:r$, $i=1:m$}
        	\State $\mathbf{h}_i^q(s) \gets \sum_{j \in \mathcal{N}_i} w_{ij}\mathbf{h}_j^{q-1}(s)$ \Comment{Consensus rounds}
        \EndFor
        \For{$i=1:m$}
        	\State $\mathbf{x}_i(s+1) \gets P_{\mathbf{x}_i(s)}(\gamma \mathbf{h}_i^r(s))$ \Comment{Prox mapping}
            \State $\mathbf{x}_i^\mathrm{av}(s+1) \gets \frac{1}{s}\sum_{k=1}^s\mathbf{x}_i(k)$ \Comment{Average iterates}
        \EndFor
    \EndFor
  \end{algorithmic}
  \Return $\mathbf{x}_i^{\mathrm{av}}(S+1), i=1,\dots,m.$
\end{algorithm}

In D-SAMD, each node $i$ initializes its iterate at the minimizer of $\omega(\cdot)$, which is guaranteed to be unique due to strong convexity. At each mini-batch round $s$, each node $i$ obtains its mini-batched subgradient and nodes engage in $r$ rounds of averaging consensus to produce the (approximate) average subgradients $\mathbf{h}_i^r(s)$. Then, each node $i$ takes a mirror prox step, using $\mathbf{h}^r_i(s)$ instead of its own mini-batched estimate. Finally, each node keeps a running average of its iterates, which is well-known to speed up convergence \cite{Polyak.Juditsky.JCO1992}.

\subsection{Convergence Analysis}
The convergence rate of D-SAMD depends on the bias and variance of the approximate subgradient averages $\mathbf{h}_i^r(s)$. In principle, averaging subgradients together reduces the noise variance and speeds up convergence. However, because averaging consensus using only $r$ communications rounds results in {\em approximate} averages, each node takes a slightly different mirror prox step and therefore ends up with a different iterate. At each mini-batch round $s$, nodes then compute subgradients at different search points, leading to bias in the averages $\mathbf{h}^r_i(s)$. This bias accumulates at a rate that depends on the subgradient noise variance, the topology of the network, and the number of consensus rounds per mini-batch round.

Therefore, the first step in bounding the convergence speed of D-SAMD is to bound the bias and the variance of the subgradient estimates $\mathbf{h}^r_i(s)$, which we do in the following lemma.
\begin{lemma}\label{lem:consensus.error.norm}
	Let $0 \leq \lambda_2 < 1$ denote the magnitude of the second-largest (ordered by magnitude) eigenvalue of $\mathbf{W}$. Define the matrices
    \begin{align*}
    	\mathbf{H}(s) &\triangleq [\mathbf{h}_1^r(s), \dots, \mathbf{h}_m^r(s)], \\
        \mathbf{G}(s) &\triangleq [\mathbf{g}(\mathbf{x}_1(s)), \dots, \mathbf{g}(\mathbf{x}_m(s))], \text{ and} \\
        \mathbf{Z}(s) &\triangleq [\mathbf{z}_1(s), \dots, \mathbf{z}_m(s)],
    \end{align*}
    recalling that the subgradient noise $\mathbf{z}_i(s)$ is defined with respect to the mini-batched subgradient $\theta_i(s)$. Also define
    \begin{align*}
    	\overline{\mathbf{g}}(s) &\triangleq \frac{1}{m}\sum_{i=1}^m \mathbf{g}(\mathbf{x}_i(s)), \quad \overline{\mathbf{G}}(s) \triangleq [\overline{\mathbf{g}}(s), \dots, \overline{\mathbf{g}}(s)], \text{ and}\\
        \overline{\mathbf{z}}(s) &\triangleq \frac{1}{m}\sum_{i=1}^m\mathbf{z}_i(s), \quad \overline{\mathbf{Z}}(s)  \triangleq [\overline{\mathbf{z}}(s), \cdots, \overline{\mathbf{z}}(s)],
    \end{align*}
    where the matrices $\overline{\mathbf{G}}(s), \overline{\mathbf{Z}}(s) \in \mathbb{R}^{n \times n}$ have identical columns. Finally, define the matrices
    \begin{align*}
    	\mathbf{E}(s) &\triangleq \mathbf{G}(s)\mathbf{W}^r - \overline{\mathbf{G}}(s) \text{ and} \\
        \tilde{\mathbf{Z}}(s) &\triangleq \mathbf{Z}(s)\mathbf{W}^r - \overline{\mathbf{Z}}(s)
    \end{align*}
    of average consensus error on the subgradients and subgradient noise, respectively. Then, the following facts are true. First, one can write $\mathbf{H}(s)$ as
    \begin{equation}\label{eqn:gradient.decomposition}
    	\mathbf{H}(s) = \overline{\mathbf{G}}(s) + \mathbf{E}(s) + \overline{\mathbf{Z}}(s) + \tilde{\mathbf{Z}}(s).
    \end{equation}
    Second, the columns of $\overline{\mathbf{Z}}(s)$ satisfy
    \begin{equation}
    	E[\norm{\overline{\mathbf{z}}(s)}_*^2] \leq \frac{C_*\sigma^2}{mb}.
    \end{equation}
    Finally,  the $i$th columns of $\mathbf{E}(s)$ and $\tilde{\mathbf{Z}}(s)$, denoted by $\mathbf{e}_i(s)$ and $\tilde{\mathbf{z}}_i(s)$, respectively, satisfy
    \begin{equation}\label{eqn:gradient.average.norm}
    	\norm{\mathbf{e}_i(s)}_* \leq \max_{j,k} m^2\sqrt{C_*}\lambda_2^r \norm{\mathbf{g}_j(s) - \mathbf{g}_k(s)}_*
    \end{equation}
    and
    \begin{equation}\label{eqn:average.gradient.noise.variance}
    	E[\norm{\tilde{\mathbf{z}}_i(s)}_*^2] \leq \frac{\lambda^{2r}m^2 C_* \sigma^2}{b},
    \end{equation}
    where we have used $\mathbf{g}_j(s)$ as a shorthand for $\mathbf{g}(\mathbf{x}_j(s))$.
\end{lemma}

The next step in the convergence analysis is to bound the distance between iterates at different nodes. As long as iterates are not too far apart, the subgradients computed at different nodes have sufficiently similar means that averaging them together reduces the overall subgradient noise.
\begin{lemma}\label{lem:iterate.gap}
	Let $a_s \triangleq \max_{i,j} \norm{\mathbf{x}_i(s) - \mathbf{x}_j(s)}$. The moments of $a_s$ follow:
    \begin{align}
    	E[a_s] &\leq \frac{\mathcal{M}+\sigma/\sqrt{b}}{L}((1+\alpha m^2 \sqrt{C_*} \lambda_2^r)^{s}-1), \\
        E[a_s^2] &\leq \frac{(\mathcal{M}+\sigma/\sqrt{b})^2}{L^2}((1+\alpha m^2 \sqrt{C_*} \lambda_2^r)^{s}-1)^2.
    \end{align}
\end{lemma}

Now, we bound D-SAMD's expected gap to optimality.
\begin{theorem}\label{thm:mirror.descent.convergence.rate}
	For D-SAMD, the expected gap to optimality at each node $i$ satisfies
    \begin{multline}\label{eqn:DSAMD.convergence.rate}
    	E[\psi(\mathbf{x}_i^{\mathrm{av}}(S+1))] - \psi^* \leq \\ \frac{2L\Omega_\omega^2}{\alpha S} + \sqrt{\frac{2(4\mathcal{M}^2 + 2\Delta_S^2)}{\alpha S}} +  \sqrt{\frac{\alpha}{2}}\frac{\Xi_S D_\omega}{L},
    \end{multline}
where
\begin{align}
	\Xi_s &\triangleq \left(\mathcal{M}+\frac{\sigma}{\sqrt{b}}\right)(1+ m^2 \sqrt{C_*} \lambda_2^r)\times\nonumber\\
&\qquad\qquad\qquad ((1+\alpha m^2 \sqrt{C_*} \lambda_2^r)^{s}-1) + 2\mathcal{M}
\end{align}
and
\begin{align}
    \Delta_s^2 &\triangleq 2 \left(\mathcal{M}+\frac{\sigma}{\sqrt{b}}\right)^2(1+m^4 C_* \lambda_2^{2r})\times\nonumber\\
    &\qquad\qquad((1+\alpha m^2 \sqrt{C_*} \lambda_2^r)^{s}-1)^2 + 4C_*\sigma^2/(mb) \nonumber\\
    &\qquad\qquad\qquad+ 4\lambda_2^{2r}C_*\sigma^2 m^2/b +4\mathcal{M}
\end{align}
quantify the moments of the effective subgradient noise.
\end{theorem}

The convergence rate proven in Theorem \ref{thm:mirror.descent.convergence.rate} is akin to that provided in \cite{Lan.MP12}, with $\Delta_s^2$ taking the role of the subgradient noise variance. A crucial difference is the presence of the final term involving $\Xi_s$. In \cite{Lan.MP12}, this term vanishes because the intrinsic subgradient noise has zero mean. However, the equivalent gradient error in D-SAMD does not have zero mean in general. As nodes' iterates diverge, their subgradients differ, and the nonlinear mapping between iterates and subgradients results in noise with nonzero mean.

The critical question is how fast communication needs to be for order-optimum convergence speed, i.e., the convergence speed that one would obtain if nodes had access to other nodes' subgradient estimates at each round. After $S$ mini-batch rounds, the network has processed $mT$ data samples. Centralized mirror descent, with access to all $mT$ data samples in sequence, achieves the convergence rate \cite{Lan.MP12}
\begin{equation*}
	O(1)\left[\frac{L}{mT} + \frac{\mathcal{M} + \sigma}{\sqrt{mT}} \right].
\end{equation*}
The final term dominates the error as a function of $m$ and $T$ if $\sigma^2  > 0$. In the following corollary we derive conditions under which the convergence rate of D-SAMD matches this term.
\begin{corollary}\label{cor:mirror.descent.consensus.rounds}
	The optimality gap for D-SAMD satisfies
     \begin{equation}
        E[\psi(\mathbf{x}_i^\mathrm{av}(S+1))] - \psi^* = O\left(\frac{\mathcal{M} + \sigma}{\sqrt{mT}} \right),
    \end{equation}
    provided the mini-batch size $b$, the communications ratio $\rho$, the number of users $m$, and the Lipschitz constant $\mathcal{M}$ satisfy
    \begin{align*}
    	b &= \Omega\left(1 + \frac{\log(mT)}{\rho\log(1/\lambda_2)}\right), \quad b = O\left(\frac{\sigma T^{1/2}}{m^{1/2}}\right),\\
        \rho &= \Omega\left(\frac{m^{1/2}\log(mT)}{\sigma T^{1/2}\log(1/\lambda_2)}\right), \quad T = \Omega\left(\frac{m}{\sigma^2}\right), \text{ and}\\
        \mathcal{M} &= O\left(\min\left\{\frac{1}{m},\frac{1}{\sqrt{ m \sigma^2 T}}\right\}\right).
    \end{align*}
\end{corollary}

\subsection{Discussion}
Corollary \ref{cor:mirror.descent.consensus.rounds} gives new insights into influences of the communications and streaming rates, network topology, and mini-batch size on the convergence rate of distributed stochastic learning. In \cite{Dekel.etal.JMLR2012}, a mini-batch size of $b=O(T^{1/2})$ is prescribed---which is sufficient whenever gradient averages are perfect---and in \cite{tsianos.rabbat.SIPN2016} the number of imperfect consensus rounds needed to facilitate the mini-batch size $b$ prescribed in \cite{Dekel.etal.JMLR2012} is derived. By contrast, we derive a mini-batch condition sufficient to drive the effective noise variance to $O(\sigma^2/(mT))$ while taking into consideration the impact of imperfect subgradient averaging. This condition depends not only on $T$ but also on $m$, $\rho$, $\lambda_2$, and $\sigma^2$---indeed, for all else constant, the optimum mini-batch size is merely $\Omega(\log(T))$. Then, the condition on $\rho$ essentially ensures that $b = O(T^{1/2})$ as specified in \cite{Dekel.etal.JMLR2012}.

We note that Corollary \ref{cor:mirror.descent.consensus.rounds} imposes a strict requirement on $\mathcal{M}$, the Lipschitz constant of the non-smooth part of $\psi$. Essentially the non-smooth part must vanish as $m$, $T$, or $\sigma^2$ becomes large. This is because the contribution of $h(\mathbf{x})$ to the convergence rate depends only on the number of iterations taken, not on the noise variance. Reducing the effective subgradient noise via mini-batching has no impact on this contribution, so we require the Lipschitz constant $\mathcal{M}$ to be small to compensate.

Finally, we note that Corollary \ref{cor:mirror.descent.consensus.rounds} dictates the relationship between the size of the network and the number of data samples obtained at each node. Leaving the terms besides $m$ and $T$ constant, Corollary \ref{cor:mirror.descent.consensus.rounds} requires $T = \Omega(m)$, i.e. the number of nodes in the network should scale no faster than the number of data samples processed per node. This is a relatively mild condition for big data applications; many applications involve data streams that are large relative to the size of the network. Furthermore, ignoring the $\log(mT)$ term and assuming $\lambda_2$ and $\sigma$ to be fixed, Corollary \ref{cor:mirror.descent.consensus.rounds} indicates that a communication ratio of $\rho = \Omega\big(\sqrt{m/T}\big)$ is sufficient for order optimality; i.e., nodes need to communicate at least $\Omega\big(\sqrt{m/T}\big)$ times per data sample. This means that if $T$ scales faster than $\Omega(m)$ then the required communications ratio approaches zero in this case as $m,T \to \infty$. In particular, fast stochastic learning is possible in expander graphs, for which the spectral gap $1-\lambda_2$ is bounded away from zero, even in communication rate-limited scenarios. For graph families that are poor expanders, however, the required communications ratio depends on the scaling of $\lambda_2$ as a function of $m$.

\section{Accelerated Distributed Stochastic Approximation Mirror Descent}\label{sect:accelerated.mirror.descent}
In this section, we present {\em accelerated} distributed stochastic approximation mirror descent (AD-SAMD), which distributes the accelerated stochastic approximation mirror descent proposed in \cite{Lan.MP12}. The centralized version of accelerated mirror descent achieves the optimum convergence rate of
\begin{equation*}
	O(1)\left[\frac{L}{T^2}+\frac{\mathcal{M} + \sigma^2}{\sqrt{T}} \right].
\end{equation*}
Consequently, we will see that the convergence rate of AD-SAMD has $1/S^2$ as its first term. This faster convergence in $S$ allows for more aggressive mini-batching, and the resulting conditions for order-optimal convergence are less stringent.

\subsection{Description of AD-SAMD}
The setting for AD-SAMD is the same as in Section \ref{sect:mirror.descent}. We again suppose a distance function $\omega: X \to \mathbb{R}$, its associated prox function/Bregman divergence $V: X \times X \to \mathbb{R}$, and the resulting (Lipschitz) prox mapping $P_x: \mathbb{R}^n \to X$.

As in Section \ref{sect:dsamd.description}, we suppose a mixing matrix $\mathbf{W} \in \mathbb{R}^{m \times m}$ that is symmetric, doubly stochastic, consistent with $G$, and has nonzero spectral gap. The main distinction between accelerated and standard mirror descent is the way one averages iterates. Rather than simply average the sequence of iterates, one maintains several distinct sequences of iterates, carefully averaging them along the way. This involves two sequences of step sizes $\beta_s \in [1,\infty)$ and $\gamma_s \in \mathbb{R}$, which are not held constant. Again we suppose that the number of mini-batch rounds $S=T/b$ is predetermined. We detail the steps of AD-SAMD in Algorithm \ref{alg:accelerated}.

\begin{algorithm}[t]
  \caption{Accelerated distributed stochastic approximation mirror descent (AD-SAMD)
    \label{alg:accelerated}}
    \begin{algorithmic}[1]
    \Require Doubly-stochastic matrix $\mathbf{W}$, step size sequences $\gamma_s$, $\beta_s$, number of consensus rounds $r$, batch size $b$, and stream of mini-batched subgradients $\theta_i(s)$.
    \For{$i=1:m$}
    	\State $\mathbf{x}_i(1),\mathbf{x}^\mathrm{md}_i(1),\mathbf{x}^\mathrm{ag}_i(1) \gets \min_{\mathbf{x} \in X} \omega(\mathbf{x})$ \Comment{Initialize search points}
    \EndFor
    \For {$s=1:S$}
        \For {$i=1:m$}
        	\State $\mathbf{x}_i^\mathrm{md}(s) \gets \beta_s^{-1}\mathbf{x}_i(s) + (1-\beta^{-1}_s)\mathbf{x}_i^\mathrm{ag}(s)$
            \State $\mathbf{h}_i^0(s) \gets \theta_i(s)$ \Comment{Get mini-batched subgradients}
        \EndFor

        \For{$q=1:r$, $i=1:m$}
        	\State $\mathbf{h}_i^q(s) \gets \sum_{j \in \mathcal{N}_i} w_{ij}\mathbf{h}_j^{q-1}(s)$ \Comment{Consensus rounds}
        \EndFor
        \For{$i=1:m$}
        	\State $\mathbf{x}_i(s+1) \gets P_{\mathbf{x}_i(s)}(\gamma_s \mathbf{h}_i^r(s))$ \Comment{Prox mapping}
            \State $\mathbf{x}^\mathrm{ag}_i(s+1) \gets \beta_s^{-1}\mathbf{x}_i(s+1) + (1-\beta_s^{-1})\mathbf{x}_i^\mathrm{ag}(s)$
        \EndFor
    \EndFor
  \end{algorithmic}
  \Return $\mathbf{x}_i^{\mathrm{ag}}(S+1), i=1,\dots,m.$
\end{algorithm}

The sequences of iterates $\mathbf{x}_i(s)$, $\mathbf{x}_i^{\mathrm{md}}(s)$, and $\mathbf{x}^\mathrm{ag}_i(s)$ are interrelated in complicated ways; we refer the reader to \cite{Lan.MP12} for an intuitive explanation of these iterations.

\subsection{Convergence Analysis}
As with D-SAMD, the convergence analysis relies on bounds on the bias and variance of the averaged subgradients. To this end, we note first that Lemma \ref{lem:consensus.error.norm} also holds for AD-SAMD, where $\mathbf{H}(s)$ has columns corresponding to noisy subgradients evaluated at $\mathbf{x}_i^\mathrm{md}(s)$. Next, we bound the distance between iterates at different nodes. This analysis is somewhat more complicated due to the relationships between the three iterate sequences.
\begin{lemma}\label{lem:accelerated.iterate.gap}
	Let
    \begin{align*}
    	a_s &\triangleq \max_{i,j}\norm{\mathbf{x}^\mathrm{ag}_i(s) - \mathbf{x}^\mathrm{ag}_j(s)}, \\
        b_s &\triangleq \max_{i,j}\norm{\mathbf{x}_i(s) - \mathbf{x}_j(s)}, \text{ and} \\
        c_s &\triangleq \max_{i,j}\norm{\mathbf{x}^\mathrm{md}_i(s) - \mathbf{x}^\mathrm{md}_j(s)}.
    \end{align*}
    Then, the moments of $a_s$, $b_s$, and $c_s$ satisfy:
    \begin{align*}
    	E[a_s],E[b_s],E[c_s] &\leq \frac{\mathcal{M} \!\!+\! \sigma/\sqrt{b}}{L}((1 \!\!+\! 2\gamma_s m^2 \sqrt{C_*}L\lambda_2^r)^s \!-\! 1), \\
        E[a_s^2],E[b_s^2],E[c_s^2] &\leq \frac{(\mathcal{M} \!\!+\! \sigma/\sqrt{b})^2}{L^2}(\!(1 \!\!+\! 2\gamma_s m^2 \!\!\sqrt{C_*} L\lambda_2^r)^s \!\!-\!\! 1)^2.
    \end{align*}
\end{lemma}

Now, we bound the expected gap to optimality of the AD-SAMD iterates.
\begin{theorem}\label{thm:accelerated.optimality.gap}
	For AD-SAMD, there exist step size sequences $\beta_s$ and $\gamma_s$ such that the expected gap to optimality satisfies
    \begin{multline}
    	E[\Psi(\mathbf{x}_i^\mathrm{ag}(S+1))] - \Psi^* \leq \frac{8 L D_{\omega,X}^2}{\alpha S^2} + \\ 4 D_{\omega,X}\sqrt{\frac{4M + \Delta_S^2}{\alpha S}} + \sqrt{\frac{32}{\alpha}}D_{\omega,X}\Xi_S,
    \end{multline}
    where
    \begin{multline*}
    	\Delta_\tau^2 = 2(\mathcal{M}+\sigma/\sqrt{b})^2((1+ 2\gamma_\tau m^2 \sqrt{C_*} L\lambda_2^r)^\tau-1)^2 + \\ \frac{4 C_* \sigma^2}{b}(\lambda_2^{2r}m^2 + 1/m) + 4\mathcal{M}.
     \end{multline*}
     and
     \begin{multline*}
        \Xi_\tau = (\mathcal{M} + \sigma/\sqrt{b})(1+\sqrt{C_*}m^2\lambda_2^r) \times \\ ((1+2\gamma_\tau m^2\sqrt{C_*}L\lambda_2^r)^\tau-1) + 2\mathcal{M}.%
    \end{multline*}
\end{theorem}

As with D-SAMD, we study the conditions under which AD-SAMD achieves order-optimum convergence speed. The centralized version of accelerated mirror descent, after processing the $mT$ data samples that the network sees after $S$ mini-batch rounds, achieves the convergence rate
\begin{equation*}
	O(1)\left[ \frac{L}{(mT)^2} + \frac{\mathcal{M}+\sigma}{\sqrt{mT}}\right].
\end{equation*}
This is the optimum convergence rate under any circumstance. In the following corollary, we derive the conditions under which the convergence rate matches the second term, which usually dominates the error when $\sigma^2 > 0$.
\begin{corollary}\label{cor:accelerated.descent.consensus.rounds}
	The optimality gap satisfies
    \begin{equation*}
    	E[\psi(\mathbf{x}^\mathrm{ag}_i(S+1)] - \psi^* = O\left(\frac{\mathcal{M} + \sigma}{\sqrt{mT}} \right),
    \end{equation*}
    provided
    \begin{align*}
    	b &= \Omega\left(1 + \frac{\log(mT)}{\rho\log(1/\lambda_2)}\right), \quad b = O\left(\frac{\sigma^{1/2}T^{3/4}}{m^{1/4}}\right), \\
        \rho &= \Omega\left( \frac{m^{1/4}\log(m T)}{\sigma T^{3/4}\log(1/\lambda_2)} \right), \quad T = \Omega\left(\frac{m^{1/3}}{\sigma^2}\right), \text{ and}\\
        \mathcal{M} &= O\left(\min\left\{\frac{1}{m},\frac{1}{\sqrt{ m \sigma^2 T}}\right\}\right).
    \end{align*}
\end{corollary}

\subsection{Discussion}
The crucial difference between the two schemes is that AD-SAMD has a convergence rate of $1/S^2$ in the absence of noise and non-smoothness. This faster term, which is often negligible in centralized mirror descent, means that AD-SAMD tolerates more aggressive mini-batching without impact on the order of the convergence rate. As a result, while the condition on the mini-batch size $b$ is the same in terms of $\rho$, the condition on $\rho$ is relaxed by $1/4$ in the exponents of $m$ and $T$. This is because the condition $b = O(T^{1/2})$, which holds for standard stochastic SO methods, is relaxed to $b = O(T^{3/4})$ for accelerated stochastic mirror descent.

Similar to Corollary \ref{cor:mirror.descent.consensus.rounds}, Corollary \ref{cor:accelerated.descent.consensus.rounds} prescribes a relationship between $m$ and $T$, but the relationship for AD-SAMD is $T~=~\Omega(m^{1/3})$, holding all but $m,T$ constant. This again is due to the relaxed mini-batch condition $b = O(T^{3/4})$ for accelerated stochastic mirror descent. Furthermore, ignoring the $\log$ term, Corollary \ref{cor:accelerated.descent.consensus.rounds} indicates that a communications ratio $\rho = \Omega\left(\frac{m^{1/4}}{T^{3/4}}\right)$ is needed for well-connected graphs such as expander graphs. In this case, as long as $T$ grows faster than the cube root of $m$, order-optimum convergence rates can be obtained even for small communications ratio. Thus, the use of accelerated methods increases the domain in which order optimum rate-limited learning is guaranteed.

\section{Numerical Example: Logistic Regression}\label{sect:numerical}
To demonstrate the scaling laws predicted by Corollaries \ref{cor:mirror.descent.consensus.rounds} and \ref{cor:accelerated.descent.consensus.rounds} and to investigate the empirical performance of D-SAMD and AD-SAMD, we consider supervised learning via binary logistic regression. Specifically, we assume each node observes a stream of pairs $\xi_i(t) = (y(t),l(t))$ of
data points $y_i(t) \in \mathbb{R}^d$ and their labels $l_i(t) \in \{0,1\}$, from which it learns a classifier with the log-likelihood function
\begin{equation*}
	F(\mathbf{x},x_0,\mathbf{y},l) = l (\mathbf{y}^T\mathbf{x} + x_0) - \log(1+\exp(\mathbf{y}^T\mathbf{x} + x_0))
\end{equation*}
where $\mathbf{x} \in \mathbb{R}^d$ and $x_0 \in \mathbb{R}$ are regression coefficients.

The SO task is to learn the optimum regression coefficients $\mathbf{x},x_0$. In terms of the framework of this paper, $\Upsilon = (\mathbb{R}^d \times \{0,1\})$, and $X = \mathbb{R}^{d+1}$ (i.e., $n = d+1$). We use the Euclidean norm, inner product, and distance-generating function to compute the prox mapping. The convex objective function is the negative of the log-likelihood function, averaged over the unknown distribution of the data, i.e.,
\begin{equation*}
	\psi(\mathbf{x}) = -E_{\mathbf{y},l}[F(\mathbf{x},x_0,\mathbf{y},l)].
\end{equation*}
Minimizing $\psi$ is equivalent to performing maximum likelihood estimation of the regression coefficients \cite{bishop:book06}.

We examine performance on synthetic data so that there exists a ``ground truth'' distribution with which to compute $\psi(\mathbf{x})$ and evaluate empirical performance. We suppose that the data follow a Gaussian distribution. For $l_i(t)~\in~\{0,1\}$, we let $\mathbf{y}_i(t)~\sim~\mathcal{N}(\mu_{l_i(t)},\sigma_r^2\mathbf{I})$, where $\mu_{l(t)}$ is one of two mean vectors, and $\sigma_r^2 > 0$ is the noise variance.\footnote{The variance $\sigma_r^2$ is distinct from the resulting gradient noise variance $\sigma^2$.} For this experiment, we draw the elements $\mu_0$ and $\mu_1$ randomly from the standard normal distribution, let $d=20$, and choose $\sigma_r^2=2$. We consider several network topologies, as detailed in the next subsections.

We compare the performance of D-SAMD and AD-SAMD against several other schemes. As a best-case scenario, we consider {\em centralized} mirror descent, meaning that at each data-acquisition round $t$ all $m$ data samples and their associated gradients are available at a single machine, which carries out stochastic mirror descent and {\em accelerated} stochastic mirror descent. These algorithms naturally have the best average performance. As a baseline, we consider {\em local} (accelerated) stochastic mirror descent, in which nodes simply perform mirror descent on their own data streams without collaboration. This scheme does benefit from an insensitivity to the communications ratio $\rho$, and no mini-batching is required, but it represents a minimum standard for performance in the sense that it does not require collaboration among nodes.

Finally, we consider a communications-constrained adaptation of {\em distributed gradient descent} (DGD), introduced in \cite{Nedic.Ozdaglar.ITAC2009}, where local subgradient updates are followed by a single round of consensus averaging on the search points $\mathbf{x}_i(s)$. DGD implicitly supposes that $\rho=1$. To handle the $\rho < 1$ case, we consider two adaptations: {\em naive} DGD, in which data samples that arrive between communications rounds are simply discarded, and {\em mini-batched} DGD, in which nodes compute {\em local} mini-batches of size $b=1/\rho$, take gradient updates with the local mini-batch, and carry out a consensus round. While it is not designed for the communications rate-limited scenario, DGD has good performance in general, so it represents a natural alternative against which to compare the performance of D-SAMD and AD-SAMD.

\subsection{Fully Connected Graphs}
First, we consider the simple case of a fully connected graph, in which $E$ is the set of all possible edges, and the obvious mixing matrix is $\mathbf{W}= \mathbf{1}\mathbf{1}^T/n$, which has $\lambda_2 = 0$. This represents a best-case scenario in which to validate the theoretical claims made above. We choose $\rho = 1/2$ to examine the regime of low communications ratio, and we let $m$ and $T$ grow according to two regimes: $T = m$, and $T = \sqrt{m}$, which are the regimes in which D-SAMD and AD-SAMD are predicted to give order-optimum performance, respectively. The constraint on mini-batch size per Corollaries \ref{cor:mirror.descent.consensus.rounds} and \ref{cor:accelerated.descent.consensus.rounds} is trivial, so we take $b=2$ to ensure that nodes can average each mini-batch gradient via (perfect) consensus. We select the following step-size parameter $\gamma$: $0.5$ and $2$ for (local and centralized) stochastic mirror descent (MD) and accelerated stochastic mirror descent (A-MD), respectively; 5 for both variants of DGD; and $5$ and $20$ for D-SAMD and AD-SAMD, respectively.\footnote{While the accelerated variant of stochastic mirror descent makes use of two
sequences of step sizes, $\beta_s$ and $\gamma_s$, these two sequences can be expressed as a function of a single parameter $\gamma$; see, e.g., the proof of Theorem~\ref{thm:accelerated.optimality.gap}.} These values were selected via trial-and-error to give good performance for all algorithms; future work involves the use of adaptive step size rules such as AdaGrad and ADAM~\cite{Duchi:JMLR2011,Kingma:ICLR2015}.

\begin{figure}[htb]
  \centering
  \begin{subfigure}{0.4\textwidth}
    \includegraphics[width=\textwidth]{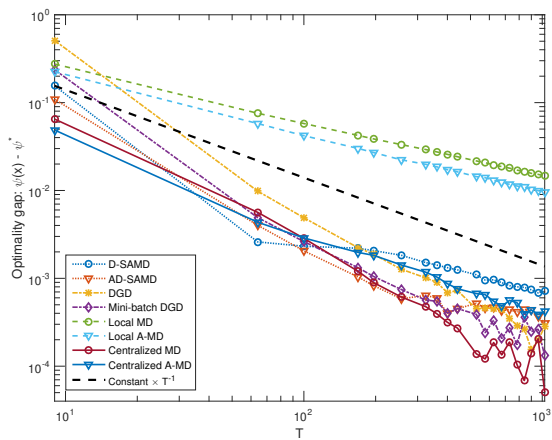}
   \caption{$T = m$}
   \end{subfigure}
     \begin{subfigure}{0.4\textwidth}
    \includegraphics[width=\textwidth]{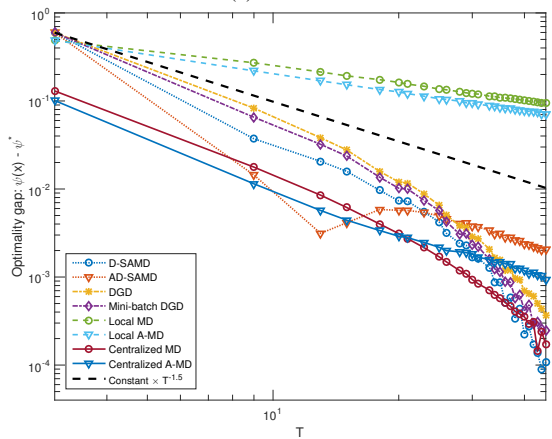}
    \caption{$T = \sqrt{m}$}
   \end{subfigure}
   \caption{Performance of different schemes for a fully connected graph on $\log$-$\log$ scale. The dashed lines (without markers) in (a) and (b) correspond to the asymptotic performance upper bounds for D-SAMD and AD-SAMD predicted by the theoretical analysis.}
	\label{fig:fully.connected}
\end{figure}

In Figure~\ref{fig:fully.connected}(a) and Figure~\ref{fig:fully.connected}(b), we plot the performance averaged over 1200 and 2400 independent instances of the problem, respectively. We also plot the order-wise theoretical performance $1/\sqrt{mT}$, which has a constant slope on the log-log axis. As expected, the distributed methods significantly outperform the local methods. The performance of the distributed methods is on par with the asymptotic theoretical predictions, as seen by the slope of the performance curves, with the possible exception of D-SAMD for $T = m$. However, we observe that D-SAMD performance is at least as good as predicted by theory for $T = \sqrt{m}$, a regime in which optimality is not guaranteed for D-SAMD. This suggests the possibility that the requirement that $T = \Omega(m)$ for D-SAMD is an artifact of the analysis, at least for this problem.

\subsection{Expander Graphs}
For a more realistic setting, we consider {\em expander graphs}, which are families of graphs that have spectral gap $1-\lambda_2$ bounded away from zero as $m \to \infty$. In particular, we use 6-regular graphs, i.e., regular graphs in which each node has six neighbors, drawn uniformly from the ensemble of such graphs. Because the spectral gap is bounded away from zero for expander graphs, one can more easily examine whether performance of D-SAMD and AD-SAMD agrees with the ideal scaling laws discussed in Corollaries \ref{cor:mirror.descent.consensus.rounds} and \ref{cor:accelerated.descent.consensus.rounds}. At the same time, because D-SAMD and AD-SAMD make use of imperfect averaging, expander graphs also allow us to examine non-asymptotic behavior of the two schemes. Per Corollaries \ref{cor:mirror.descent.consensus.rounds} and \ref{cor:accelerated.descent.consensus.rounds}, we choose $b = \frac{1}{10}\frac{\log(mT)}{\rho \log(1/\lambda_2)}$. While this choice is guaranteed to be sufficient for optimum asymptotic performance, we chose the multiplicative constant $1/10$ via trial-and-error to give good non-asymptotic performance.

\begin{figure}[htb]
  \centering
  \begin{subfigure}{0.4\textwidth}
    \includegraphics[width=\textwidth]{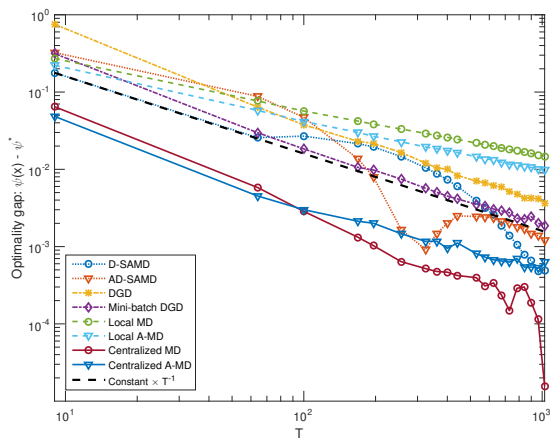}
   \caption{$T = m$}
   \end{subfigure}
     \begin{subfigure}{0.4\textwidth}
    \includegraphics[width=\textwidth]{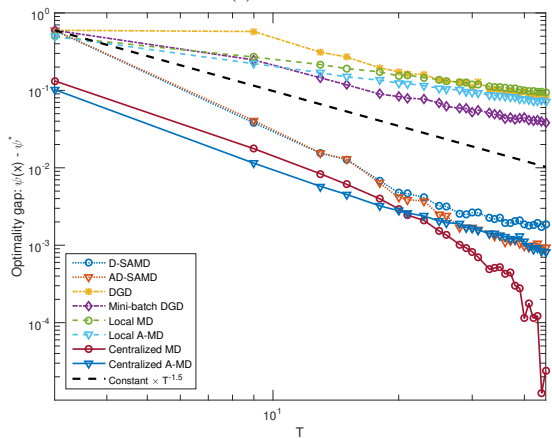}
    \caption{$T = \sqrt{m}$}
   \end{subfigure}
   \caption{Performance of different schemes for $6$-regular expander graphs on $\log$-$\log$ scale, for $\rho = 1/2$. Dashed lines once again represent asymptotic theoretical upper bounds on performance.}
	\label{fig:expander}
\end{figure}

In Figure \ref{fig:expander} we plot the performance averaged over 600 problem instances. We again take $\rho = 1/2$, and consider the regimes $T = m$ and $T = \sqrt{m}$. The step sizes are the same as in the previous subsection except that $\gamma = 2.5$ for D-SAMD when $T = \sqrt{m}$, $\gamma = 28$ for AD-SAMD when $T = m$, and $\gamma= 8$ for AD-SAMD when $T = \sqrt{m}$. Again, we see that AD-SAMD and D-SAMD outperform local methods, while their performance is roughly in line with asymptotic theoretical predictions. The performance of DGD, on the other hand, depends on the regime: For $T = m$, it appears to have order-optimum performance, whereas for $T = \sqrt{m}$ it has suboptimum performance on par with local methods. The reason for the dependency on regime is not immediately clear and suggests the need for further study into DGD-style methods in the case of rate-limited networks.

\subsection{Erd\H{o}s-Renyi Graphs}
Finally, we consider {\em Erd\H{o}s-Renyi} graphs, in which a random fraction (in this case $0.1$) of possible edges are chosen. These graphs are not expanders, and their spectral gaps are not bounded. Therefore, order-optimum performance is not easy to guarantee, since the conditions on the rate and the size of the network depend on $\lambda_2$, which is not guaranteed to be well behaved. We again take $\rho = 1/2$, consider the regimes $T = m$ and $T = \sqrt{m}$, and again we choose $b = \frac{1}{10}\frac{\log(mT)}{\rho \log(1/\lambda_2)}$. The step sizes are chosen to be the same as for expander graphs in both regimes.

\begin{figure}[htb]
  \centering
  \begin{subfigure}{0.4\textwidth}
    \includegraphics[width=\textwidth]{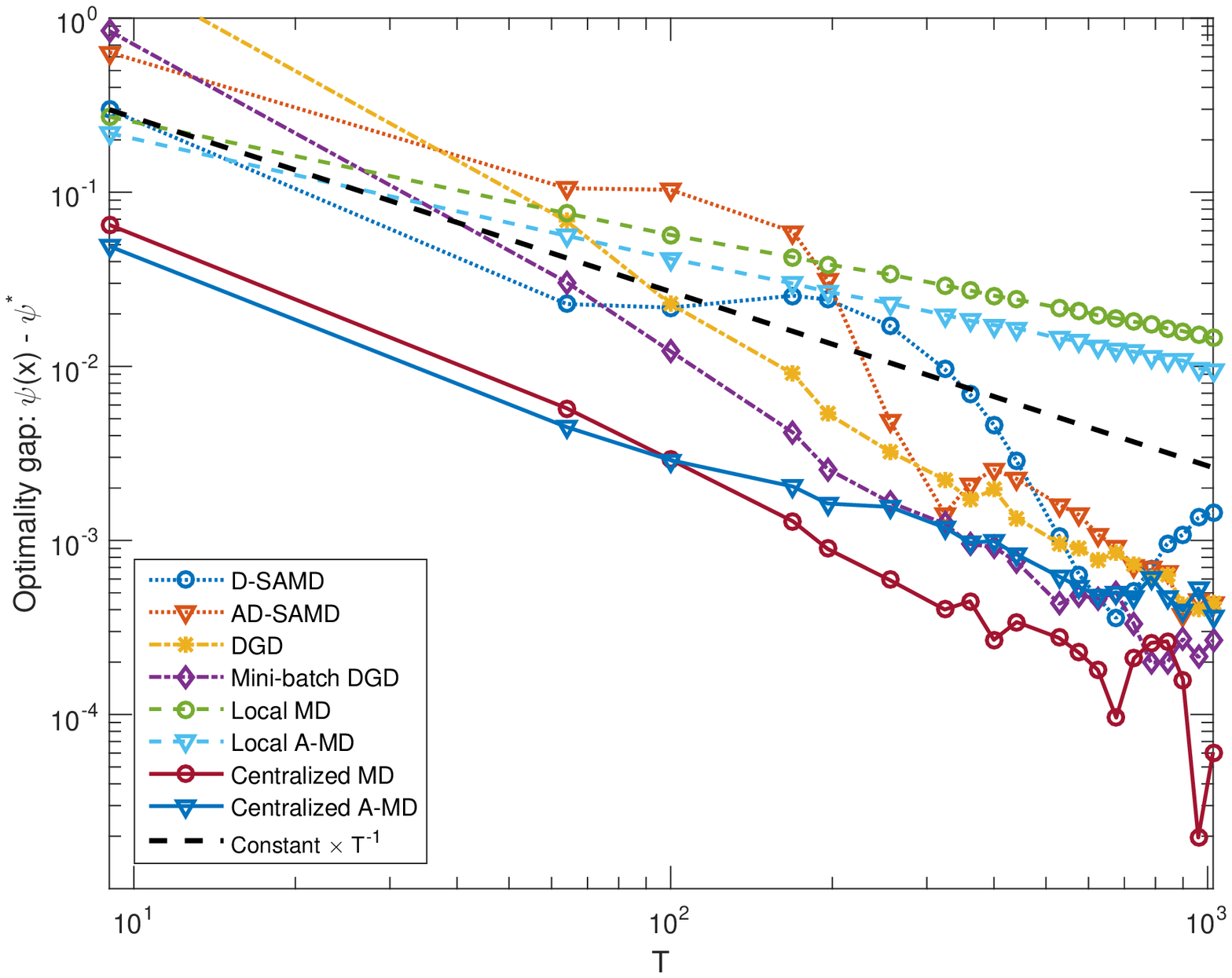}
   \caption{$T = m$}
   \end{subfigure}
     \begin{subfigure}{0.4\textwidth}
    \includegraphics[width=\textwidth]{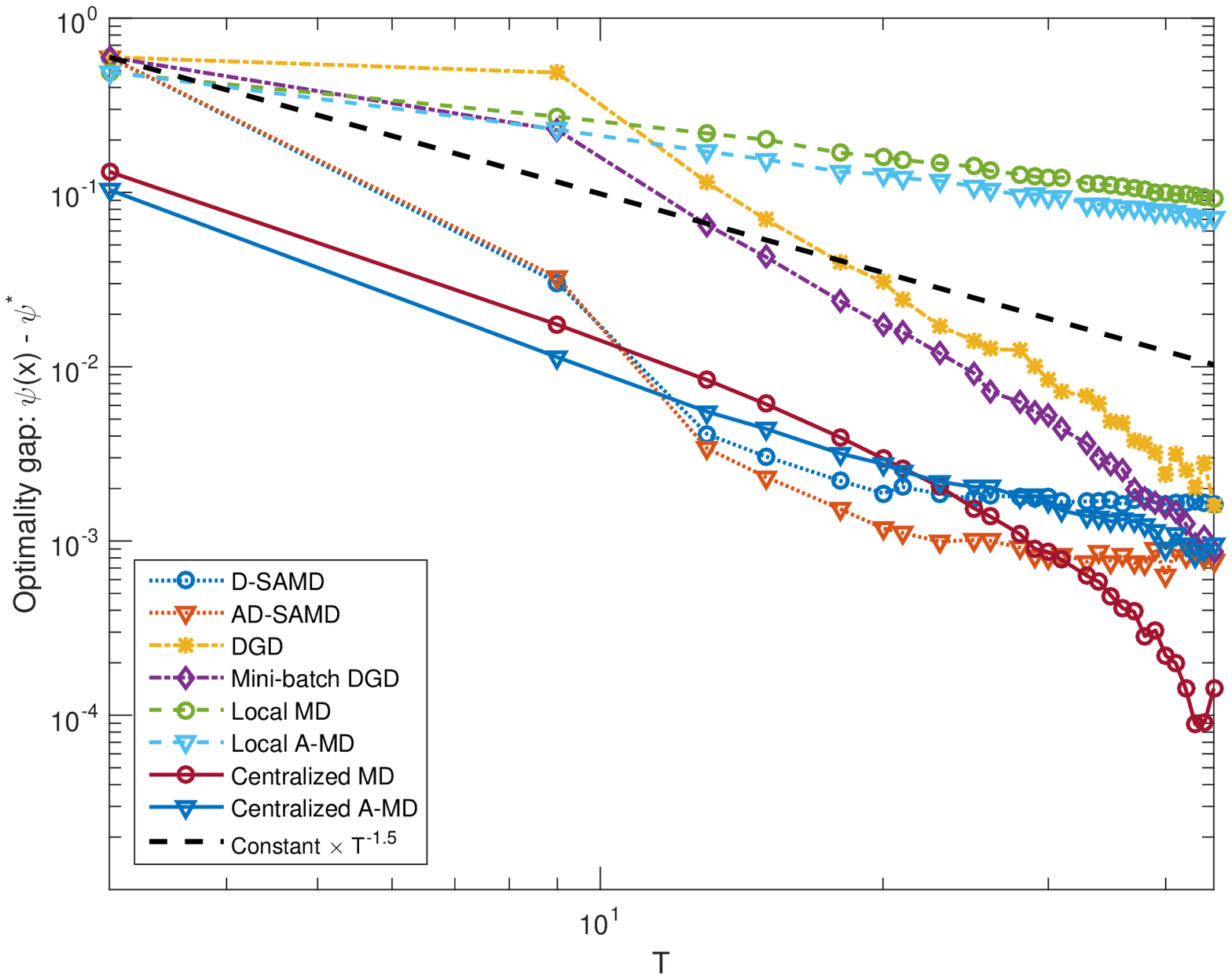}
    \caption{$T = \sqrt{m}$}
   \end{subfigure}
   \caption{Performance of different schemes on Erd\H{o}s-Renyi graphs, for $\rho = 1/2$, displayed using $\log$-$\log$ scale.}
	\label{fig:erdos}
\end{figure}

Once again, we observe a clear distinction in performance between local and distributed methods; in particular, all distributed methods (including DGD) appear to show near-optimum performance in both regimes. However, as expected the performance is somewhat more volatile than in the case of expander graphs, especially for the case of $T = m$, and it is possible that the trends seen in these plots will change as $T$ and $m$ increase.

\section{Conclusion}\label{sect:conclusion}
We have presented two distributed schemes, D-SAMD and AD-SAMD, for convex stochastic optimization over networks of nodes that collaborate via rate-limited links. Further, we have derived sufficient conditions for the order-optimum convergence of D-SAMD and AD-SAMD, showing that accelerated mirror descent provides a foundation for distributed SO that better tolerates slow communications links. These results characterize relationships between network communications speed and the convergence speed of stochastic optimization.

A limitation of this work is that we are restricted to settings in which the prox mapping is Lipschitz continuous, which excludes important Bregman divergences such as the Kullbeck-Liebler divergence. Further, the conditions for optimum convergence restrict the Lipschitz constant of non-smooth component of the objective function to be small. Future work in this direction includes study of the limits on convergence speed for more general divergences and composite objective functions.

\appendix
\subsection{Proof of Lemma \ref{lem:consensus.error.norm}}
	The first claim follows immediately from the definitions of the constituent matrices. The second claim follows from Lemma \ref{lem:average.variance}. To establish the final claim, we first bound the norm of the columns of $\mathbf{E}(s)$:
    \begin{align}
    	\mathbf{E}(s) &\triangleq \mathbf{G}(s)\mathbf{W}^r - \overline{\mathbf{G}}(s)\nonumber \\
        &= \mathbf{G}(s)(\mathbf{W}^r - \mathbf{1}\mathbf{1}^T/n),\nonumber \\
        &= (\mathbf{G}(s) - \overline{\mathbf{G}}(s))(\mathbf{W}^r - \mathbf{1}\mathbf{1}^T/n),
    \end{align}
    where the first equality follows from $\overline{\mathbf{G}}(s) = \mathbf{G}(s) \mathbf{1}\mathbf{1}^T/n$ by definition, and the second equality follows from the fact that the dominant eigenspace of a column-stochastic matrix $\mathbf{W}$ is the subspace of constant vectors, thus the rows of $\bar{\mathbf{G}}(s)$ lie in the null space of $\mathbf{W}^r - \mathbf{1}\mathbf{1}^T/n$. We bound the norm of the columns of $\mathbf{E}(s)$ via the Frobenius norm of the entire matrix:
    \begin{align*}
    	\norm{\mathbf{e}_i(s)}_* 
        &\leq \sqrt{C_1}\norm{\mathbf{E}(s)}_F \\
        &= \sqrt{C_1}\norm{(\mathbf{G}(s) - \overline{\mathbf{G}}(s))(\mathbf{W}^r - \mathbf{1}\mathbf{1}^T/n)}_F \\
        &\leq \sqrt{C_1}m\lambda_2^r\norm{\mathbf{G}(s) - \overline{\mathbf{G}}(s)}_F \\
        &\leq \max_{j,k} m^2\sqrt{C_*}\lambda_2^r \norm{\mathbf{g}_j(s) - \mathbf{g}_k(s)}_*,
    \end{align*}
    where $C_1$ bounds the gap between the Euclidean and dual norms, $C_2$ bounds the gap in the opposite direction, and thus $C_* \leq C_1C_2$.
    Then, we bound the variance of the subgradient noise. Similar to the case of $\mathbf{E}(s)$, we can rewrite
    \begin{equation*}
    	\tilde{\mathbf{Z}}(s) = \mathbf{Z}(s)(\mathbf{W}^r- \mathbf{1}\mathbf{1}^T/n).
    \end{equation*}
    We bound the expected squared norm of each column $\tilde{\mathbf{z}}_i(s)$ in terms of the expected Frobenius norm:
    \begin{align*}
    	E[\norm{\mathbf{\tilde{z}}_i(s)}_*^2] 
    	&\leq C_1 E[\norm{\mathbf{Z}(s)(\mathbf{W}^r- \mathbf{1}\mathbf{1}^T/n)}^2_F] \\
        &\leq m\lambda_2^{2r} C_1 E[\norm{\mathbf{Z}(s)}_F^2] \\
        &\leq \lambda_2^{2r} m^2 C_*  \sigma^2/b.
    \end{align*}

\subsection{Proof of Lemma \ref{lem:iterate.gap}}
	By the Lipschitz condition on $P_{\mathbf{x}}(\mathbf{y})$,
    \begin{align*}
    	a_{s+1} &= \max_{i,j} \norm{P_{\mathbf{x}_i(s)}(\gamma\mathbf{h}_i(s)) - P_{\mathbf{x}_j(s)}(\gamma\mathbf{h}_j(s))} \\
        &\leq \max_{i,j} \norm{\mathbf{x}_i(s) - \mathbf{x}_j(s)} + \gamma\norm{\mathbf{h}_i(s) - \mathbf{h}_j(s)}_* \\
        &\leq \max_{i,j} \norm{\mathbf{x}_i(s) - \mathbf{x}_j(s)} + \frac{\alpha}{2L}\norm{\mathbf{h}_i(s) - \mathbf{h}_j(s)}_*,
	\end{align*}
    where the final inequality is due to the constraint $\gamma \leq \alpha/(2L)$. Applying Lemma \ref{lem:consensus.error.norm}, we obtain
    \begin{equation*}
        a_{s+1} \leq a_s + \max_{i,j} \frac{\alpha}{2L}\norm{\mathbf{e}_i(s) - \mathbf{e}_j(s) + \tilde{\mathbf{z}}_i(s) - \tilde{\mathbf{z}}_j(s)}_*,
    \end{equation*}
    which, applying (\ref{eqn:gradient.average.norm}), yields
    \begin{multline*}
        a_{s+1} \leq a_s + \max_{i,j} \frac{\alpha m^2 \sqrt{C_*} \lambda_2^r}{2L}\norm{\mathbf{g}_i(s) - \mathbf{g}_j(s)}_* + \\ \frac{\alpha}{2L}\norm{\tilde{\mathbf{z}}_i(s) - \tilde{\mathbf{z}}_j(s)}_*.
    \end{multline*}
    Applying (\ref{eqn:subgradient.bound}), we get
    \begin{align}
        a_{s+1} &\leq a_s + \alpha m^2 \sqrt{C_*}\lambda_2^r(a_s + \mathcal{M}/L) + \max_i \frac{\alpha}{L}\norm{\tilde{\mathbf{z}}_i(s)}_* \notag \\
        &= (1\!+\!\alpha m^2 \sqrt{C_*}\lambda_2^r) a_s \!\!+\!\! \frac{\alpha}{L}\left(m^2\sqrt{C_*}\lambda_2^r \mathcal{M} \!\!+\!\! \norm{\tilde{\mathbf{z}}_{i^*}(s)}_*\right), \label{eqn:mirror.descent.recursion}
    \end{align}
    where $i^* \in V$ is the index that maximizes $\norm{\tilde{\mathbf{z}}_i}_*$.

    The final expression (\ref{eqn:mirror.descent.recursion}) gives the recurrence relationship that governs the divergence of the iterates, and it takes the form of a standard first-order difference equation. Recalling the base case $a_1=0$, we obtain the solution
    \begin{align}\label{eqn:standard.difference.solution}
    	a_s &\leq \frac{\alpha}{L}\sum_{\tau=0}^{s-1}\left(1+\alpha m^2 \!\sqrt{C_*} \lambda_2^r\right)^{s-1-\tau}\times\nonumber\\
    &\qquad\qquad\qquad\qquad\left(m^2\!\sqrt{C_*}\lambda_2^r \mathcal{M} \!+\! \norm{\tilde{\mathbf{z}}_{i^*}(s)}_*\right).
    \end{align}
	To prove the lemma, we bound the moments of $a_s$ using the solution (\ref{eqn:standard.difference.solution}). First, we bound $E[a_s^2]$:
	\begin{multline*}
        E[a_s^2] \leq   E\Bigg[\Bigg(\frac{\alpha}{L}\sum_{\tau=0}^{s-1}(1+\alpha m^2\sqrt{C_*} \lambda_2^r)^{s-1-\tau} \times \\
       \left(m^2\sqrt{C_*}\lambda_2^r \mathcal{M} + \norm{\tilde{\mathbf{z}}_{i^*}(s)}_*\right) \Bigg)^2\Bigg],
    \end{multline*}
    which we can rewrite via a change of variables as
    \begin{multline*}
        E[a_s^2] \leq \left(\frac{\alpha}{L}\right)^2 \sum_{\tau=0}^{s-1}\sum_{q=0}^{s-1}(1+\alpha m^2\sqrt{C_*}\lambda_2^r)^{\tau+q} \times \\
        E\big[(m^2\sqrt{C_*}\lambda_2^r \mathcal{M}+\norm{\tilde{\mathbf{z}}_{i^*}(s-1-\tau)}_*) \times \\
        (m^2\sqrt{C_*}\lambda_2^r \mathcal{M}+\norm{\tilde{\mathbf{z}}_{i^*}(s-1-q)}_*) \big].
	\end{multline*}
	Then, we apply the Cauchy-Schwartz inequality and (\ref{eqn:average.gradient.noise.variance})
    to obtain
	\begin{multline*}
		E[a_s^2] \leq \lambda_2^{2r}C_*m^4(\mathcal{M} + \sigma/\sqrt{b})^2\left(\frac{\alpha}{L}\right)^2 \times \\ \sum_{\tau=0}^{s-1}(1+\alpha m^2 \sqrt{C_*} \lambda_2^r)^{\tau}\sum_{q=0}^{s-1}(1+\alpha m^2 \sqrt{C_*} \lambda_2^r)^{q}.
	\end{multline*}
	Finally, we apply the geometric sum identity and simplify:
	\begin{align*}
        E[a_s^2] \! &\leq \! \lambda_2^{2r}C_*m^4(\mathcal{M} \!+\!\! \sigma/\sqrt{b})^2\!\left(\frac{\alpha}{L}\right)^2 \times\\
        &\qquad\qquad\qquad\left( \frac{1\!-\!\!(1+\alpha m^2 \sqrt{C_*} \lambda_2^r)^{s}}{1\!-\!\! (1+\alpha m^2 \sqrt{C_*} \lambda_2^r)} \right)^2 \\
        &= \frac{(\mathcal{M}+\sigma/\sqrt{b})^2}{L^2}((1+\alpha m^2 \sqrt{C_*} \lambda_2^r)^{s}-1)^2.
    \end{align*}
	This establishes the result for $E[a_s^2]$, and the bound on $E[a_s]$ follows {\em a fortiori} from Jensen's inequality.

\subsection{Proof of Theorem \ref{thm:mirror.descent.convergence.rate}}
	The proof involves an analysis of mirror descent, with the subgradient error quantified by Lemmas \ref{lem:consensus.error.norm} and \ref{lem:iterate.gap}. As necessary, we will cite results from the proof of \cite[Theorem 1]{Lan.MP12} rather than reproduce its analysis.

	Define
    \begin{align*}
    	\boldsymbol{\delta}_i(\tau) &\triangleq \mathbf{h}_i(\tau) - \mathbf{g}(\mathbf{x}_i(\tau)) \\
        \boldsymbol{\eta}_i(\tau) &\triangleq \bar{\mathbf{g}}(\tau) + \mathbf{e}_i(\tau) - \mathbf{g}(\mathbf{x}_i(\tau)) = \boldsymbol{\delta}_i(\tau) - \bar{\mathbf{z}}(\tau) - \tilde{\mathbf{z}}_i(\tau) \\
        \zeta_i(\tau) &\triangleq \gamma \langle \boldsymbol{\delta}_i(\tau), \mathbf{x}^* - \mathbf{x}_i(\tau) \rangle.
    \end{align*}
    Applying Lemmas \ref{lem:consensus.error.norm} and \ref{lem:iterate.gap}, we bound $E[\norm{\boldsymbol{\eta}_i(\tau)}_*]$:
    \begin{align*}
    	E[\norm{\boldsymbol{\eta}_i(\tau)}_*] \!\!&=\! E\left[\norm{\overline{\mathbf{g}}(\tau) + \mathbf{e}_i(\tau) - \mathbf{g}(\mathbf{x}_i(\tau)) }_*\right] \\
        &\leq\! E[\norm{\overline{\mathbf{g}}(\tau) - \mathbf{g}(\mathbf{x}_i(\tau))}_*] + E[\norm{\mathbf{e}_i(\tau)}_*] \\
        &\leq\! LE[a_\tau] + L m^2 \sqrt{C_*} \lambda_2^r E[a_\tau] + 2\mathcal{M}\\
        &\leq\! L(1+ m^2 \sqrt{C_*} \lambda_2^r)E[a_\tau] + 2\mathcal{M}\\
        &\leq\! (\mathcal{M}\!+\!\!\sigma/\sqrt{b})(1\!+\!\! m^2 \sqrt{C_*} \lambda_2^r)\times\\
        &\qquad\qquad((1\!+\!\!\alpha m^2 \sqrt{C_*} \lambda_2^r)^{s}\!-\!\!1)\! +\!\! 2\mathcal{M}.
    \end{align*}

    Next, we bound the expected value of $\zeta_i(\tau)$:
    \begin{align*}
    	E[\zeta_i(\tau)] &= \gamma E[\langle \boldsymbol{\eta}_i(\tau) +  \bar{\mathbf{z}}(\tau) - \tilde{\mathbf{z}}_i(\tau), \mathbf{x}^* - \mathbf{x}_i(\tau) \rangle] \\
         &= \gamma E[\langle \boldsymbol{\eta}_i(\tau), \mathbf{x}^* - \mathbf{x}_i(\tau) \rangle] + \\
         & \quad\quad\quad \gamma E[\langle \bar{\mathbf{z}}(\tau) - \tilde{\mathbf{z}}_i(\tau), \mathbf{x}^* - \mathbf{x}_i(\tau) \rangle] \\
        &= \gamma E[\langle \boldsymbol{\eta}_i(\tau), \mathbf{x}^* - \mathbf{x}_i(\tau) \rangle] \\
        &\leq \gamma E[\norm{\boldsymbol{\delta}_i(\tau)}_*\norm{\mathbf{x}^* - \mathbf{x}_i(\tau)}] \\
        &\leq \gamma \Xi_\tau \Omega_\omega \\
        &\leq \sqrt{\frac{\alpha}{2}}\frac{\Xi_\tau D_\omega}{L},
    \end{align*}
    where the first inequality is due to the definition of the dual norm, and the third equality follows because $\bar{\mathbf{z}}(\tau)$ and $\tilde{\mathbf{z}}_i(\tau)$ have zero mean and are independent of $\mathbf{x}^* - \mathbf{x}_i(\tau)$.

    Next, we bound $E[\norm{\boldsymbol{\delta}_i(\tau)}^2_*]$. By Lemma \ref{lem:consensus.error.norm},
    \begin{multline*}
    	E[\norm{\boldsymbol{\delta}_i(\tau)}^2_*] \!=\! E\left[\norm{\overline{\mathbf{g}}(\tau) \!+\! \mathbf{e}_i(\tau) \!+\! \overline{\mathbf{z}}(\tau) \!+\! \tilde{\mathbf{z}}_i(\tau) \!-\! \mathbf{g}(\mathbf{x}_i(\tau)) }^2_*\right],
    \end{multline*}
    which we can bound via repeated uses of the triangle inequality:
    \begin{multline*}
    	E[\norm{\boldsymbol{\delta}_i(\tau)}^2_*] \leq 2E[\norm{\bar{g}(\tau) - g(\mathbf{x}_i(\tau))}_*^2] + \\ 2E[\norm{\mathbf{e}_i(\tau)}_*^2] +  4E[\norm{\bar{\mathbf{z}}(\tau)}_*^2 + \norm{\tilde{\mathbf{z}}_t(\tau)}_*^2].
    \end{multline*}
    Next, we apply Lemma \ref{lem:consensus.error.norm}:
    \begin{multline*}
    	E[\norm{\boldsymbol{\delta}_i(\tau)}^2_*] \leq 2L^2(1+m^4 C_* \lambda_2^{2r})E[a_\tau^2] + \\ 4C_*\sigma^2/(mb) + 4\lambda_2^{2r}C_*\sigma^2 m^2/b + 4\mathcal{M}.
    \end{multline*}
    Finally, we apply Lemma \ref{lem:iterate.gap}:
        \begin{multline*}
    	E[\norm{\boldsymbol{\delta}_i(\tau)}^2_*] \leq 2L^2(1+m^4 C_* \lambda_2^{2r})E[a_\tau^2] + \\ 4C_*\sigma^2/(mb) + 4\lambda_2^{2r}C_*\sigma^2 m^2/b +4\mathcal{M}= \Delta_\tau^2.
    \end{multline*}

    Applying the proof of Theorem 1 of \cite{Lan.MP12} to the D-SAMD iterates, we have that
    \begin{multline*}
    	S\gamma (\psi(\mathbf{x}_i^\mathrm{av}(S+1)) - \psi^*) \leq D_\omega^2 + \\ \sum_{\tau=1}^S\left[\zeta_i(\tau) + \frac{2\gamma^2}{\alpha}\left(4\mathcal{M}^2 + \norm{\boldsymbol{\delta}_i(\tau)}^2_* \right) \right].
    \end{multline*}
	We take the expectation to obtain
	\begin{multline*}
    	E[\psi(\mathbf{x}_i^\mathrm{av}(S+1))] - \psi^* \leq \frac{D_\omega^2}{S\gamma} + \frac{1}{S}\sum_{\tau=1}^S\sqrt{\frac{\alpha}{2}}\frac{\Xi_\tau D_\omega}{L} + \\ \frac{2\gamma}{\alpha S}\left(4S\mathcal{M}^2 + \sum_{\tau=1}^S \Delta_\tau^2 \right).
    \end{multline*}
    Because $\{\Xi_\tau\}$ and $\{\Delta_\tau^2\}$ are increasing sequences, we can bound the preceding by
    \begin{align*}
        E[\psi(\mathbf{x}_i^\mathrm{av}(S+1))] - \psi^* &\leq \frac{D_\omega^2}{S\gamma} + \frac{2\gamma}{\alpha}(4\mathcal{M}^2
+ \Delta_S^2)\nonumber\\
&\qquad\qquad\qquad\quad+ \sqrt{\frac{\alpha}{2}}\frac{\Xi_S D_\omega}{L}.
	\end{align*}
    Minimizing this bound over $\gamma$ in the interval $(0,\alpha/(2L)]$, we obtain the following optimum value:
    \begin{equation*}
    	\gamma^* = \min\left\{\frac{\alpha}{2L},\sqrt{\frac{\alpha D_\omega^2}{2S(4\mathcal{M}^2 + 2\Delta_S^2)}} \right\},
    \end{equation*}
	which results in the bound
    \begin{multline*}
    	E[\psi(\mathbf{x}_i^\mathrm{av}(s+1))] - \psi^* \leq \frac{2LD_\omega^2}{\alpha S} + \\ \sqrt{\frac{2(4\mathcal{M}^2 + 2\Delta_S^2)}{\alpha S}} + \sqrt{\frac{\alpha}{2}}\frac{\Xi_S D_\omega}{L}.
    \end{multline*}

\subsection{Proof of Corollary \ref{cor:mirror.descent.consensus.rounds}}
	First, we make a few simplifying approximations to $\Xi_\tau$:
    \begin{align*}
    	&\Xi_\tau\\
        &=(\mathcal{M}\!+\!\!\sigma/\sqrt{b})(1\!+\!\! m^2 \sqrt{C_*} \lambda_2^r)((1\!+\!\!\alpha m^2 \sqrt{C_*} \lambda_2^r)^{s}\!-\!\!1)\! +\!\! 2\mathcal{M} \\
        &\leq  2(\mathcal{M} + \sigma/\sqrt{b}) \sqrt{C_* m}((1+\alpha m^2\sqrt{C_*} \lambda_2^r)^{\tau}) +2\mathcal{M}\\
        &= 2(\mathcal{M} + \sigma/\sqrt{b}) \sqrt{C_* m}\sum_{k=0}^\tau \binom{s}{k} (\alpha m^2\sqrt{C_*} \lambda_2^r)^k) + 2\mathcal{M}\\
        &\leq 2(\mathcal{M} + \sigma/\sqrt{b}) \sqrt{C_* m}\sum_{k=0}^\tau (\tau\alpha m^2\sqrt{C_*} \lambda_2^r)^k) + 2\mathcal{M}\\
        &\leq 2(\mathcal{M} + \sigma/\sqrt{b}) \sqrt{C_* m} (\tau+1)(\tau\alpha m^2\sqrt{C_*} \lambda_2^r) ) + 2\mathcal{M}\\
        &= O(\tau^2 \sqrt{m^5} \lambda_2^r(\mathcal{M} + \sigma/\sqrt{b}) + \mathcal{M}), \label{eqn:xi.approximation}
    \end{align*}
    where the first inequality is trivial, the next two statements are due to the binomial theorem and the exponential upper bound on the binomial coefficient, and the final inequality is true if we constrain $S\alpha m^2\sqrt{C_*} \lambda_2^r \leq 1$, which is consistent with the order-wise constraints listed in the statement of this corollary. Recalling that $r \leq b \rho$, we obtain
    \begin{equation}
    	\Xi_S = O\left(S^2 \lambda_2^{b\rho}\sqrt{m^5}(\mathcal{M} + \sigma/\sqrt{b}) + \mathcal{M}^2\right).
    \end{equation}
    Now, we find the optimum scaling law on the mini-batch size $b$. To ensure $S\alpha m^2\sqrt{C_*} \lambda_2^r \leq 1$ for all $s$, we need
    \begin{equation}
        b = \Omega\left( \frac{\log(mT)}{\rho \log(1/\lambda_2)}\right).
    \end{equation}
    In order to get optimum scaling we need $\Xi_S~=~O(\sigma/\sqrt{mT} + \mathcal{M})$, which yields a slightly stronger necessary condition:
    \begin{equation}\label{eqn:optimum.minibatch}
    	b = \Omega\left(1+\frac{\log(mT)}{\rho \log(1/\lambda_2)}\right),
    \end{equation}
    where the first term is necessary because $b$ cannot be smaller than unity.

    Similar approximations establish that
    \begin{equation*}
    	\Delta^2_S = O(S^4 m^5 \lambda_2^{2b\rho}(\mathcal{M}+\sigma^2/b) + \sigma^2/(mb) + \lambda_2^{2b\rho}\sigma^2 m^2 /b + \mathcal{M}^2).
    \end{equation*}
    Thus, (\ref{eqn:optimum.minibatch}) ensures $\Delta_S^2 = O(\sigma^2/(mb) + \mathcal{M}^2)$. The resulting gap to optimality scales as
	\begin{equation*}
    	E[\Psi(\mathbf{x}_i^\mathrm{av}(S+1))] - \Psi^* = O\left(\frac{b}{T} + \sqrt{\frac{b\mathcal{M}^2}{T}} + \frac{\sigma}{\sqrt{mT}} +\mathcal{M} \right).
    \end{equation*}
    Substituting (\ref{eqn:optimum.minibatch}) yields
    \begin{multline*}
        E[\Psi(\mathbf{x}_i^\mathrm{av}(S+1))] - \Psi^* = O\Bigg(\max\left\{\frac{\log(mT)}{\rho T\log(1/\lambda_2)},\frac{1}{T}\right\} + \\ \max \left\{\sqrt{\frac{\mathcal{M}^2\log(mT)}{\rho T\log(1/\lambda_2)}},\sqrt{\frac{\mathcal{M}^2}{T}}\right\} + \frac{\sigma}{\sqrt{mT}} +\mathcal{M} \Bigg).
    \end{multline*}
    In order to achieve order optimality, we need the first term to be $O((\mathcal{M}+\sigma)/\sqrt{mT})$, which requires
    \begin{equation*}
    	b = O\left(\frac{\sigma T^{1/2}}{m^{1/2}}\right), \ m = O(\sigma^2 T), \ \rho = \Omega\left( \frac{m^{1/2}\log(m T)}{\sigma T^{1/2}\log(1/\lambda_2)}\right).
    \end{equation*}
    Finally, we also need the second and fourth terms to be $O((\mathcal{M}~+~\sigma)/\sqrt{mT})$, which requires
\begin{equation}\label{eqn:dsamd.m.orderwise.condition}
    	\mathcal{M} = O\left(\min\left\{\frac{1}{m},\frac{1}{\sqrt{ m \sigma^2 T}}\right\}\right).
    \end{equation}
    This establishes the result.

\subsection{Proof of Lemma \ref{lem:accelerated.iterate.gap}}
	The sequences $a_s$, $b_s$, $c_s$ are interdependent. Rather than solving the complex recursion directly, we bound  the dominant sequence. Define $d_s \triangleq \max\{a_s,b_s,c_s\}$. First, we bound $a_s$:
    \begin{multline*}
    	a_{s+1} = \max_{i,j} ||\beta_{s+1}^{-1}(\mathbf{x}_i(s+1) - \mathbf{x}_j(s+1) + \\
        (1-\beta_{s+1}^{-1})(\mathbf{x}_i^\mathrm{ag}(s) - \mathbf{x}_j^{\mathrm{ag}(s)})||,
    \end{multline*}
    which, via the triangle inequality, is bounded by
    \begin{multline*}
    	a_{s+1} \leq \max_{i,j} \beta_{s+1}^{-1}\norm{P_{\mathbf{x}_i(s)}(\gamma_s\mathbf{h}_i(s)) - P_{\mathbf{x}_j(s)}(\gamma_s\mathbf{h}_j(s))} + \\ (1-\beta_{s+1}^{-1})a_s.
    \end{multline*}
    Because the prox-mapping is Lipschitz,
    \begin{multline*}
    	a_{s+1} \leq \max_{i,j} \beta_{s+1}^{-1}(\norm{\mathbf{x}_i(s) - \mathbf{x}_j(s)} + \gamma_s\norm{\mathbf{h}_i(s) - \mathbf{h}_j(s)}_*) \\ + (1-\beta_{s+1}^{-1})a_s.
    \end{multline*}
    Applying the first part of Lemma \ref{lem:consensus.error.norm} yields
    \begin{multline*}
    	a_{s+1} \leq \max_{i,j} \beta_{s+1}^{-1}b_s + (1-\beta_{s+1}^{-1})a_s + \\ \beta_{s+1}^{-1}\gamma_s\norm{\mathbf{e}_i(s) - \mathbf{e}_j(s) + \tilde{\mathbf{z}}_i(s) - \tilde{\mathbf{z}}_j(s)}_*,
    \end{multline*}
    and applying the second part and the triangle inequality yields
    \begin{multline*}
    	a_{s+1} \leq \max_{i,j} d_s + 2\gamma_sm^2\sqrt{C_*}\lambda_2^r\norm{\mathbf{g}_i(s) - \mathbf{g}_j(s)}_* + \\ 2\gamma_s\norm{\tilde{\mathbf{z}}_i(s)}_*.
    \end{multline*}
    We apply (\ref{eqn:subgradient.bound}) and collect terms to obtain
    \begin{align*}
        a_{s+1} &\leq \max_{i} d_s + 2\gamma_sm^2\sqrt{C_*} \lambda_2^r (Lc_s + 2\mathcal{M}) + 2\gamma_s\norm{\tilde{\mathbf{z}}_i(s)} \\
        &\leq \max_i (1 + 2\gamma_sm^2\sqrt{C_*} L \lambda_2^r)d_s + \\
        & \quad\quad\quad 2\gamma_s(2m^2\sqrt{C_*}\lambda_2^r\mathcal{M} + \norm{\tilde{\mathbf{z}}_i(s)}_*).
    \end{align*}
    Next, we bound $b_s$ via similar steps:
    \begin{align*}
    	b_{s+1} &= \max_{i,j} \norm{P_{\mathbf{x}_i(s)}(\gamma_s\mathbf{h}_i(s)) - P_{\mathbf{x}_j(s)}(\gamma_s\mathbf{h}_j(s))} \\
        &\leq \max_{i,j} \norm{\mathbf{x}_i(s) - \mathbf{x}_j(s)} + \gamma_s\norm{\mathbf{h}_i(s) - \mathbf{h}_j(s)}_* \\
        &\leq \max_{i} b_s \!+\! 2\gamma_s(m^2 \sqrt{C_*}\lambda_2^r(Lc_s + 2\mathcal{M}) \!+\! \norm{\tilde{\mathbf{z}}_i(s)}_*) \\
        &\leq \max_i (1 + 2\gamma_sm^2\sqrt{C_*} L \lambda_2^r)d_s + \\
        & \quad\quad\quad 2\gamma_s(2m^2\sqrt{C_*}\lambda_2^r\mathcal{M} + \norm{\tilde{\mathbf{z}}_i(s)}_*).
    \end{align*}
    Finally, we bound $c_s$:
    \begin{multline*}
    	c_{s+1} = \max_{i,j} ||\beta_s^{-1}(\mathbf{x}_i(s+1) - \mathbf{x}_j(s+1)) + \\
        (1-\beta_s^{-1})(\mathbf{x}_i^\mathrm{ag}(s) - \mathbf{x}_j^{\mathrm{ag}}(s))||.
    \end{multline*}
    We bound this sequence in terms of $a_s$ and $b_s$:
    \begin{align*}
        c_{s+1} &\leq \max_{i,j} \beta_s^{-1}b_{s+1} + (1-\beta_s^{-1})a_s \\
        &\leq \max_i (1 + 2\gamma_sm^2\sqrt{C_*} L \lambda_2^r)d_s + \\
        & \quad\quad\quad 2\gamma_s(2m^2\sqrt{C_*}\lambda_2^r\mathcal{M} + \norm{\tilde{\mathbf{z}}_i(s)}_*).
    \end{align*}
    Comparing the three bounds, we observe that
    \begin{multline*}
    	d_{s+1} \leq \max_i (1 + 2\gamma_sm^2\sqrt{C_*} L \lambda_2^r)d_s + \\
        2\gamma_s(2m^2\sqrt{C_*}\lambda_2^r\mathcal{M} + \norm{\tilde{\mathbf{z}}_i(s)}_*).
    \end{multline*}
    Solving the recursion yields
    \begin{align}
    	d_s &\leq \sum_{\tau=0}^{s-1}(1+ 2 m^2 \gamma_s\sqrt{C_*} L \lambda_2^r)^\tau 2\gamma_s \times \nonumber \\
    &\qquad\qquad\qquad\qquad(2m^2\sqrt{C_*}\lambda_2^r\mathcal{M} + \norm{\tilde{\mathbf{z}}_i(s)}_*).
    \end{align}
    Following the same argument as in Lemma \ref{lem:iterate.gap}, we obtain
    \begin{align*}
    	E[d_s^2]
        &\leq \frac{(\mathcal{M} + \sigma/\sqrt{b})^2}{L^2}((1+ 2\gamma_s m^2 \sqrt{C_*} L\lambda_2^r)^s-1)^2.
    \end{align*}
	The bound on $E[d_s]$ follows {\em a fortiori}.

\subsection{Proof of Theorem \ref{thm:accelerated.optimality.gap}}
	Similar to the proof of Theorem \ref{thm:mirror.descent.convergence.rate}, the proof involves an analysis of accelerated mirror descent, and as necessary we will cite results from \cite[Theorem 2]{Lan.MP12}. Define
    \begin{align*}
    	\boldsymbol{\delta}_i(\tau) &= \mathbf{h}_i(\tau) - g(\mathbf{x}_i^\mathrm{md}(\tau)) \\
        \boldsymbol{\eta}_i(\tau) &\triangleq \bar{\mathbf{g}}(\tau) + \mathbf{e}_i(\tau) - \mathbf{g}(\mathbf{x}^\mathrm{md}_i(\tau)) = \boldsymbol{\delta}_i(\tau) - \bar{\mathbf{z}}(\tau) - \tilde{\mathbf{z}}_i(\tau) \\
        \zeta_i(\tau) &= \gamma_\tau \langle \delta_i(\tau), \mathbf{x}^* - \mathbf{x}_i(\tau) \rangle.
    \end{align*}
    Applying (\ref{eqn:subgradient.bound}) and Lemmas \ref{lem:consensus.error.norm} and \ref{lem:accelerated.iterate.gap}, we bound $E[\norm{\boldsymbol{\eta}_i(\tau)}]$:
    \begin{align*}
    	E[\norm{\boldsymbol{\eta}_i(\tau)}_*] &= E\left[\norm{\overline{g}(\tau) + \mathbf{e}_i(\tau) - g(\mathbf{x}_i(\tau)) }_*\right] \\
        &\leq E[\norm{\overline{\mathbf{g}}(\tau) - g(\mathbf{x}^\mathrm{md}_i(\tau))}_*] + E[\norm{\mathbf{e}_i(\tau)}_*] \\
        &\leq LE[c_\tau] + L\sqrt{C_*}m^2\lambda_2^rE[c_\tau] + 2\mathcal{M}\\
        &\leq (\mathcal{M} + \sigma/\sqrt{b})(1+\sqrt{C_*}m^2\lambda_2^r)\times \\
        &\quad\quad\quad((1+2\gamma_\tau m^2\sqrt{C_*}L\lambda_2^r)^\tau-1) + 2\mathcal{M}.
    \end{align*}
    By the definition of the dual norm:
    \begin{multline*}
    	E[\zeta_i(\tau)] \leq \gamma_\tau E[\norm{\eta_i(\tau)}_*\norm{\mathbf{x}^* - \mathbf{x}_i(\tau)}] \leq \\
        \gamma_\tau \Xi_\tau \Omega_\omega \leq \sqrt{\frac{\alpha}{2}}\frac{\Xi_\tau D_\omega}{L}.
    \end{multline*}
   We bound $E[\norm{\boldsymbol{\delta}_i(\tau)}^2]$ via the triangle inequality:
   \begin{align*}
   		E[\norm{\boldsymbol{\delta}_i(\tau)}^2] \leq &2E[\norm{\bar{\mathbf{g}}(\tau) - g(\mathbf{x}_t^\mathrm{md}(\tau)) }_*^2] + 2E[\norm{\mathbf{e}_i(\tau)}_*^2]\\
        &\qquad + 4 E[\norm{\tilde{\mathbf{z}}_i(\tau)}_*^2] + 4 E[\norm{\bar{\mathbf{z}}(\tau)}_*^2],
   \end{align*}
   which by (\ref{eqn:subgradient.bound}) and Lemma \ref{lem:consensus.error.norm} is bounded by
   \begin{multline*}
   		E[\norm{\boldsymbol{\delta}_i(\tau)}^2] \leq 2L^2(1+m^4C_* \lambda_2^{2r})E[c_\tau^2] + \\ \frac{4 C_* \sigma^2}{b}(\lambda_2^{2r}m^2 + 1/m) + 4\mathcal{M}.
   \end{multline*}
   Applying Lemma \ref{lem:accelerated.iterate.gap}, we obtain
   \begin{multline*}
   		E[\norm{\boldsymbol{\delta}_i(\tau)}^2] \leq 2(\mathcal{M}+\sigma/\sqrt{b})^2((1+ 2\gamma_\tau m^2 \sqrt{C_*} L\lambda_2^r)^\tau-1)^2 + \\ \frac{4 C_* \sigma^2}{b}(\lambda_2^{2r}m^2 + 1/m) + 4\mathcal{M}.
   \end{multline*}
   From the proof of \cite[Theorem 2]{Lan.MP12}, we observe that
   \begin{multline*}
   		(\beta_{S+1}-1)\gamma_{S+1}[\Psi(\mathbf{x}_i^\mathrm{ag}(S+1)) - \Psi^*] \leq D^2{\omega,X} + \\ \sum_{\tau=1}^S\zeta_\tau + \frac{2}{\alpha}(4M^2 + \norm{\boldsymbol{\delta}_i(\tau)}_*^2)\gamma_\tau^2)].
   \end{multline*}
   Taking the expectation yields
    \begin{multline*}
        (\beta_{S+1}-1)\gamma_{S+1}[E[\Psi(\mathbf{x}_i^\mathrm{ag}(S+1))]  - \Psi^*] \leq D^2_{\omega,X} + \\ \sqrt{\frac{2}{\alpha}}D_{\omega,X}\sum_{\tau=1}^S \gamma_\tau \Xi_\tau + \frac{2}{\alpha}\sum_{\tau=1}^S \gamma_\tau^2(4M^2 + \Delta_\tau^2).
   \end{multline*}
   Letting $\beta_\tau = \frac{\tau+1}{2}$ and $\gamma_\tau = \frac{\tau+1}{2}\gamma$, observing that $\Delta_\tau$ and $\Xi_\tau$ are increasing in $\tau$, and simplifying, we obtain
   \begin{multline*}
   		E[\Psi(\mathbf{x}_i^\mathrm{ag}(S+1))]  - \Psi^* \leq \frac{4D_{\omega,X}^2}{(S^2+1)\gamma} + \\ \sqrt{\frac{32}{\alpha}}D_{\omega,X}\Xi_S + \frac{4\gamma S}{\alpha}(4M^2 + \Delta_S^2).
   \end{multline*}
   Solving for the optimum $\gamma$ in the interval $0 \leq \gamma \leq \alpha/(2L)$, we obtain:
   \begin{equation*}
   		\gamma^* = \min\left\{\frac{\alpha}{2L}, \sqrt{\frac{ \alpha  D_{\omega,X}^2}{S(S^2+1)(4M^2 + \Delta_S^2)}} \right\}.
   \end{equation*}
   This gives the bound
   \begin{multline*}
   		E[\Psi(\mathbf{x}_i^\mathrm{ag}(S+1))]  - \Psi^* \leq \frac{8 L D_{\omega,X}^2}{\alpha(S^2+1)} + \\ \frac{16 S D_{\omega,X}^2(4M^2 + \Delta_S^2)}{\alpha(S+1)^2} + \sqrt{\frac{32}{\alpha}}D_{\omega,X}\Xi_S,
   \end{multline*}
   which simplifies to the desired bound
   \begin{multline*}
        E[\Psi(\mathbf{x}_i^\mathrm{ag}(S+1))]  - \Psi^* \leq \frac{8 L D_{\omega,X}^2}{\alpha S^2} + \\ 4 D_{\omega,X}\sqrt{\frac{4M + \Delta_S^2}{\alpha S}} + \sqrt{\frac{32}{\alpha}}D_{\omega,X}\Xi_S.
   \end{multline*}

\subsection{Proof of Corollary \ref{cor:accelerated.descent.consensus.rounds}}
	First, we bound $\Xi_\tau$ and $\Delta_\tau^2$ order-wise. Using steps similar to those in the proof of Corollary \ref{cor:mirror.descent.consensus.rounds}, one can show that
    \begin{equation*}
    	\Xi_\tau = O\left( \tau^3 m^3 \lambda_2^r(\mathcal{M} + \sigma/\sqrt{b}) + \mathcal{M} \right),
    \end{equation*}
    and
    \begin{equation*}
    	\Delta_\tau^2 = O\left(\tau^6 m^6 \lambda_2^{2r}(\mathcal{M}^2 + \sigma^2/b) + \frac{\sigma^2}{mb} + \mathcal{M}^2 \right).
    \end{equation*}
    In order to achieve order-optimum convergence, we need $\Delta_S^2~=~O(\sigma^2/(mb) + \mathcal{M}^2)$. Combining the above equation with the constraint $r \leq b\rho$, we obtain the condition
    \begin{equation*}
    	b = \Omega\left(1 +  \frac{\log(m T)}{\rho \log(1/\lambda_2)} \right).
    \end{equation*}
    This condition also ensures that $\Xi_S = O(\sigma/\sqrt{mT} + \mathcal{M})$, as is necessary for optimum convergence speed. This leads to a convergence speed of
	\begin{equation*}
    	E[\Psi(\mathbf{x}_i^\mathrm{av}(S+1))] - \Psi^* = O\left(\frac{b^2}{T^2} + \sqrt{\frac{b\mathcal{M}^2}{T}} + \frac{\sigma}{\sqrt{mT}} +\mathcal{M} \right).
    \end{equation*}
    In order to make the first term $O((\mathcal{M}+\sigma)/\sqrt{mT})$, we need
    \begin{align*}
    	b &= O\left(\frac{\sigma^{1/2}T^{3/4}}{m^{1/4}}\right), \quad m=O(\sigma^2T), \text{ and}\\
        \rho &= \Omega\left( \frac{m^{1/4}\log(m T)}{\sigma T^{3/4}\log(1/\lambda_2)} \right).
    \end{align*}
    To make the second and fourth terms $O((\mathcal{M}+\sigma)/\sqrt{mT})$, we need, as before,
    \begin{equation*}
    	\mathcal{M} = O\left(\min\left\{\frac{1}{m},\frac{1}{\sqrt{ m \sigma^2 T}}\right\}\right).
    \end{equation*}


\begin{IEEEbiography}[{\includegraphics[width=1in,height=1.25in,clip,keepaspectratio]{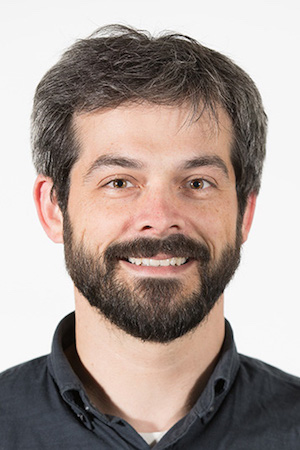}}]
{Matthew Nokleby}(S'04--M'13) received the B.S. (cum laude) and M.S. degrees from Brigham Young University, Provo, UT, in 2006 and 2008, respectively, and the Ph.D. degree from Rice University, Houston, TX, in 2012, all in electrical engineering. From 2012--2015 he was a postdoctoral research associate in the Department of Electrical and Computer Engineering at Duke University, Durham, NC. In 2015 he joined the Department of Electrical and Computer Engineering at Wayne State University, where he is an assistant professor. His research interests span machine learning, signal processing, and information theory, including distributed learning and optimization, sensor networks, and wireless communication. Dr. Nokleby received the Texas Instruments Distinguished Fellowship (2008-2012) and the Best Dissertation Award (2012) from the Department of Electrical and Computer Engineering at Rice University.
\end{IEEEbiography}

\begin{IEEEbiography}[{\includegraphics[width=1in,height=1.25in,clip,keepaspectratio]{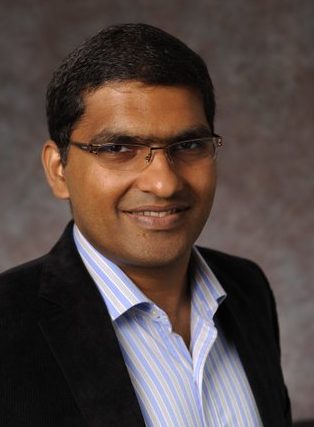}}]
{Waheed U. Bajwa}(S'98--M'09--SM'13) received BE (with Honors) degree in electrical engineering from the National University of Sciences and Technology, Pakistan in 2001, and MS and PhD degrees in electrical engineering from the University of Wisconsin-Madison in 2005 and 2009, respectively. He was a Postdoctoral Research Associate in the Program in Applied and Computational Mathematics at Princeton University from 2009 to 2010, and a Research Scientist in the Department of Electrical and Computer Engineering at Duke University from 2010 to 2011. He has been with Rutgers University since 2011, where he is currently an associate professor in the Department of Electrical and Computer Engineering and an associate member of the graduate faculty of the Department of Statistics and Biostatistics. His research interests include statistical signal processing, high-dimensional statistics, machine learning, harmonic analysis, inverse problems, and networked systems.

Dr. Bajwa has received a number of awards in his career including the Best in Academics Gold Medal and President's Gold Medal in Electrical Engineering from the National University of Sciences and Technology (2001), the Morgridge Distinguished Graduate Fellowship from the University of Wisconsin-Madison (2003), the Army Research Office Young Investigator Award (2014), the National Science Foundation CAREER Award (2015), Rutgers University's Presidential Merit Award (2016), Rutgers Engineering Governing Council ECE Professor of the Year Award (2016, 2017), and Rutgers University's Presidential Fellowship for Teaching Excellence (2017). He is a co-investigator on the work that received the Cancer Institute of New Jersey's Gallo Award for Scientific Excellence in 2017, a co-author on papers that received Best Student Paper Awards at IEEE IVMSP 2016 and IEEE CAMSAP 2017 workshops, and a Member of the Class of 2015 National Academy of Engineering Frontiers of Engineering Education Symposium. He served as an Associate Editor of the IEEE Signal Processing Letters (2014--2017), co-guest edited a special issue of Elsevier Physical Communication Journal on ``Compressive Sensing in Communications'' (2012), co-chaired CPSWeek 2013 Workshop on Signal Processing Advances in Sensor Networks and IEEE GlobalSIP 2013 Symposium on New Sensing and Statistical Inference Methods, and served as the Publicity and Publications Chair of IEEE CAMSAP 2015, General Chair of the 2017 DIMACS Workshop on Distributed Optimization, Information Processing, and Learning, and a Technical Co-Chair of the IEEE SPAWC 2018 Workshop. He is currently serving as a Senior Area Editor for IEEE Signal Processing Letters, an Associate Editor for IEEE Transactions on Signal and Information Processing over Networks, and a member of MLSP, SAM, and SPCOM Technical Committees of IEEE Signal Processing Society.
\end{IEEEbiography}
\end{document}